\PassOptionsToPackage{sort&compress}{natbib}
\documentclass[final,3p,12pt]{elsarticle}
\makeatletter
\def\ps@pprintTitle{%
  \let\@oddhead\@empty
  \let\@evenhead\@empty
  \def\@oddfoot{\reset@font\hfil\thepage\hfil}
  \let\@evenfoot\@oddfoot
}
\makeatother
\usepackage{amssymb}
\usepackage{tabularray}
\usepackage{placeins}

\usepackage{booktabs}
\usepackage{amsmath}
\usepackage{lineno}
\usepackage{makecell} % Include the makecell package
\usepackage{siunitx} % For \SI and units
\usepackage{xcolor}
\usepackage{graphicx}
\usepackage{subfigure}
\usepackage{setspace}
\usepackage{array}
\begin{document}

\begin{frontmatter}

\title{Integrating Multi-Physics Simulations and Machine Learning to Define the Spatter Mechanism and Process Window in Laser Powder Bed Fusion}

\author[inst1]{Olabode T. Ajenifujah}
\affiliation[inst1]{organization={Department of Mechanical Engineering, Carnegie Mellon University},%Department and Organization
            %addressline={5000 Forbes Avenue}, 
            city={Pittsburgh},
            postcode={15213}, 
            state={PA},
            country={USA}}
            
\author[inst1]{Francis Ogoke}
\author[inst2]{Florian Wirth}

\affiliation[inst2]{organization={Exentis Group AG},%Department and Organization
            %addressline={ Im Stetterfeld 1}, 
            city={Stetten},
            postcode={5608}, 
            country={Switzerland}}

\author[inst1]{Jack Beuth}

\author[inst1,inst3,inst4]{Amir Barati Farimani\corref{corauthor}}
\affiliation[inst3]{organization={Department of Chemical Engineering, Carnegie Mellon University},%Department and Organization
            city={Pittsburgh},
            postcode={15213}, 
            state={PA},
            country={USA}}
\affiliation[inst4]{organization={Machine Learning Department, Carnegie Mellon University},%Department and Organization
            city={Pittsburgh},
            postcode={15213}, 
            state={PA},
            country={USA}}

\begin{abstract}

Spatter is a primary source of defects in laser powder bed fusion (LPBF), yet predicting where and when it forms across a broad process space remains difficult: high-fidelity simulations that resolve spatter physics are accurate but far too computationally expensive to map a wide range of process parameters. This work introduces a framework that couples high-fidelity multi-physics simulation, machine learning, and a commercial AM-specialized solver (here, FLOW-3D) to make high-fidelity spatter prediction both transferable and physically interpretable, and uses it to generate a process map at a fraction of the usual cost. The main contributions are: (1) a transferable machine learning approach in which a classifier trained on high-fidelity OpenFOAM simulations, which resolve the three-dimensional melt pool and recoil-driven spatter ejection, is applied to melt-pool fields from a computationally efficient commercial AM solver, enabling high-fidelity spatter prediction without high-fidelity cost; (2) a spatter process map spanning 281 power--velocity conditions, built roughly 18 times faster than with high-fidelity simulation alone (approximately 4 versus 72 hours per case); and (3) a mechanistic account of spatter formation obtained through feature-importance and SHAP analyses, which identify pressure, vertical velocity, and the temperature-dependent balance between recoil pressure and surface tension as dominant drivers of ejection. The high-fidelity model is validated against experimental measurements of melt-pool dimensions, temperature, and spatter diameter and velocity, and the classifier achieves 94\% accuracy on held-out data. The resulting map links laser power and scan speed to predicted spatter behavior and identifies operating windows with minimal defects, offering a fast, physically grounded route to process-window selection in metal AM.

\end{abstract}


\end{frontmatter}


\section{Introduction}

\label{sec:sample1}
Additive manufacturing (AM), often referred to as 3D printing, is a transformative approach to industrial production. Laser powder bed fusion (LPBF) is the most common metal AM process. It has several advantages in comparison to conventional manufacturing technologies like milling and casting. LPBF offers design flexibility, quick fabrication of prototypes directly from digital models, reduced material waste, and demonstrates onsite manufacturing capability.  The process of using LPBF to create a 3D object starts with spreading a layer of metal particles on the build plate. Then the laser source moving at a certain power and velocity are guided by the laser scanners according to the information in the 3D CAD to melt the regions relating to the design, which then solidifies to
form a single layer. The powder spreading, melting and solidifying processes will occur repeatedly until the final object is formed. LPBF has attracted applications in key economic sectors such as medicine, automotive, aerospace, construction, energy, and prototyping \cite{wirth2022implementation, repossini2017use}. However, compared to other forms of manufacturing such
as machining, casting, and rolling, AM currently has the lowest market share, which is due to poor repeatability, defect formation, and low productivity \cite{wirth2022implementation, cunningham2017synchrotron}. Issues with poor repeatability can be attributed to various process parameters such as scanning speed, laser power, beam radius, gas flow path, layer thickness, and hatch spacing involved in operating LPBF. Slight changes in these parameters can propagate, resulting in variability in the properties of parts. Furthermore, the multi-physics phenomena such as heat transfer,
solidification, melting, vaporization, and cooling that occur in AM further make it a sensitive process \cite{panwisawas2020metal, yin2020correlation, li2022review}.

 Defects in LPBF could result from the process parameters that generate the melt pool, and the common types are pore \cite{hojjatzadeh2020direct, martin2019dynamics, mukherjee2018mitigation, promoppatum2022quantification, ren2023machine, cunningham2017synchrotron, gordon2020defect} and crack \cite{young2020types, anwar2018study, wang2020investigation}. Taherkhani et al. \cite{taherkhani2021development} used the in-situ photoanode sensor to analyze the light signal emitted from the melt pool to quantify the defect and found it to correlate with the process condition. 
 Spatter ejected from the melt pool is known to cause defects such as surface roughness, porosity, and inclusions \cite{young2020types, anwar2018study, wang2020investigation, ly2017metal}. Snow et al. \cite{snow2023observation} demonstrated that spatter particles play a direct role in causing stochastic lack-of-fusion defects in LPBF components. Using spatial statistics, they revealed that the locations of internal flaws are intrinsically connected to the distribution of spatter particles. Toyserkani et al. \cite{esmaeilizadeh2019effect} found that spatter increased surface roughness from 14.4 $\mu m$ to 28 $\mu m$, recommending avoiding spatter-rich regions for high-quality parts. The desire to implement multi-lasers to increase printing productivity has been inhibited by spatter generation \cite{wirth2021influence}. Spatter formation results from complex interactions between high-speed vapor, vapor-induced powder movement, and melt  pool dynamics. 
The detrimental effect of spatter could be reduced by manipulating the inert gas flow within the machine.
It is challenging to establish an optimum inert gas flow rate for spatter removal, also considering that the relationship will have to be established for the different power / velocity combinations, which is not only time consuming, but also expensive \cite{yin2020correlation, barrett2018low, anwar2018study, o2024computational}. Providing solution to spatter issues will require a holistic understanding of the physics of laser-material interactions, from melt pool creation to how the material properties can influence the intrinsic nature of the melt pool, leading to the instabilities that cause defect generation across wide process condition space.

The two general approaches to investigating different phenomena in AM are advanced observational technologies and numerical simulation methods. Few experimental studies have been conducted on spatter, Barrett et al. \cite{barrett2019statistical} used a high-speed sterovision camera to capture the spatter, then used computer vision to collect their properties for statistical analysis. Previous work by Chen et al. \cite{young2020types} revealed five categories of spatter that can be formed during LPBF: solid, metallic jet, powder agglomeration, entrainment melting spatter, and induced spatter. Additionally, they found that the size, speed, and direction of the spatters vary based on the process condition. 
Although advanced monitoring techniques have been used to investigate the mechanism of spatter formation via the physical process, establishing a quantitative relationship between spatter formation and the physical process remains challenging due to experimental limitations in resolution and scale. Determining suitable processing parameters using a trial-and-error approach is time-consuming and expensive. Additionally, it is challenging to fully monitor the complex physical phenomena in the LPBF at microsecond and micrometer scales.

Numerical simulation can offer valuable insight into areas where measurements are challenging or impossible to obtain. It also provides three-dimensional, time-dependent views of keyhole and molten pool behaviors that cannot be directly observed through experiments. Deng et al. \cite{deng2019investigation} developed a model of remote laser spiral spot welding to investigate spatter formation. Their findings revealed that the opening area of the keyhole acts as a primary out-gassing channel, and the instantaneous vapor pressure on the interface is closely related to spatter formation. Many commercial simulation packages such as Ansys, Comsol multiphysics, and StarCCM+ are used for investigating AM. The available AM specific commercial software like FLOW-3D still require adjustments of various numerical parameters, which sometimes are not accessible to the end-users.  An open-source software like OpenFOAM (Open Field Operation and Manipulation), is flexible in implementing additional features to address novel problems and processes occurring during AM process \cite{msmsac4a26bib17, msmsac4a26bib8}. However, modeling AM process by taking into account the multiphysics involved in the melt pool dynamics and spatter generation is computationally expensive.

 The high computational cost associated with high-fidelity simulations has motivated their synergy with ML tools, especially where a large scale of simulation experiments is necessary, such as in understanding defect formation processes, part design optimization, and the discovery of metamaterials and process monitoring \cite{repossini2017use, ogoke2023convolutional, yang2019predictive, ogoke2023inexpensive, wang2020machine, stathatos2019real, jadhav2023stressd, sheikh2024exploring, yao2017hybrid, bendsoe2005topology, akbari2022meltpoolnet, farimani2024fast}. Hemmasian et al. \cite{hemmasian2023surrogate} utilized FLOW-3D simulation software to generate data sets for LPBF processes and trained a convolutional neural network to predict the temperature field using only the process conditions as input to the model. DebRoy et al. \cite{du2021physics} integrated physics-informed machine learning, mechanistic modeling, and experimental data to identify key variables that shed light on the physics behind defect formation in AM.

The potential for synergy of ML with physics-based simulation to understand spatter formation remains unexploited, which could have been due to limited studies and understanding of spatter as compared to other forms of defects. Spatter generation is influenced by complex and often non-linear interactions between multiple process variables. ML models are adept at capturing these complex relationships in high-dimensional data \cite{akbari2022meltpoolnet, wang2020machine}. ML can provide predictive insights that are not easily discernible through traditional physics-based modeling.  Lastly, ML can help optimize process parameters not only to avoid spatter but also to achieve the best possible results in terms of speed, cost, and material properties \cite{wang2020investigation}. Using ML models, a deeper understanding of the spatter can be achieved, leading to improved control, improved process efficiency, and superior material characteristics in manufacturing and other material processing fields. 

\begin{figure*}[htbp!]
\begin{center}
\includegraphics[width=1\linewidth]{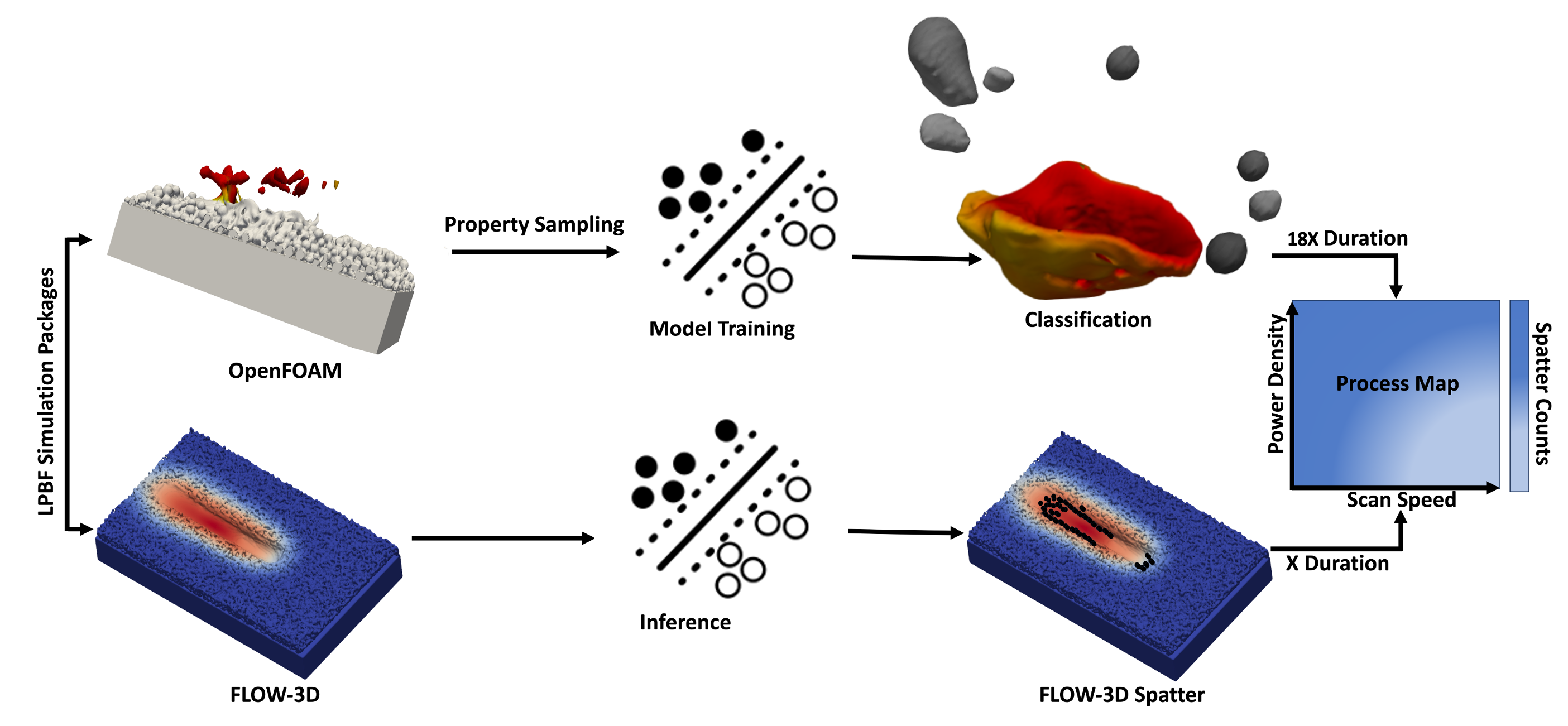}
\end{center}
    \caption{Spatter ejection and melt pool parameters from OpenFOAM simulations were used as input for machine learning classification tasks. The trained models were tested on FLOW-3D datasets, which provide computationally efficient melt-pool simulations but do not directly resolve the hot-spatter dynamics considered in this study. A 5 $\mu m$ LPBF simulation in FLOW-3D and OpenFOAM requires approximately 4 hours and 72 hours, respectively, on the same machine. }
    
\label{fig:schematic}
\end{figure*}

In this study, we develop a high fidelity multi-physics model based on OpenFOAM simulation to study the hot or droplet spatter ejection in LPBF. The formation of hot spatter has been correlated to formation of macroscopic defects, which have a  significant impact on the quality of the LPBF process. Wang et al. \cite{wang2021characteristics}  extracted features from hot spatter to train an AdaBoost-CART model for predicting single-track formation quality, achieving an accuracy of 71.1 \%.  In addition, hot spatter can cause local melting upon landing on the solid part and its solidification  can lead to formation of oxide layer, which has a higher melting point, which does not undergo melting easily acting as an impurity inside the part \cite{liu2015investigation, andani2017spatter}.  Therefore, our work aims at developing a methodology specifically for analyzing the properties that differ between the ejected fluid and the melt pool, as detecting fluctuations in the melt pool is critical for ensuring satisfactory quality control during part fabrication. Hot spatter is formed from the melt pool as a result of the force imbalance between the recoil pressure, marangoni forces, gravity and surface tension. Capturing and understanding the behavior of these spatter  will not only help distinguish them from other forms of spatter but also improve our understanding of the formation mechanisms of various spatter types.  We develop a tracking algorithm specifically for the LPBF dataset generated with OpenFOAM,  which  differentiates between spatter and the melt pool, links data across consecutive time steps, and extracting properties at each time step, such as temperature, velocity, location, and density in 3D. The scope of the present study is therefore limited to hot or droplet spatter generated directly from melt-pool instabilities in SS316L. Powder entrainment, cold spatter, agglomerated particles, and other spatter mechanisms are not explicitly modeled in the current framework and are identified as important extensions for future DEM--CFD and multi-class machine learning studies.  We unveil the relationship between the spatter ejection and the melt pool via statistical analysis, whereby spatter is detected at a short time scale (5 µs), paving the way to quantifying the fluctuation in the melt pool for good quality control. Upon coupling our ML model with multiphysics models, for the first time we reveal importance of different parameters to the spatter ejection and formulate a spatter formation mechanism. In addition to using ML models for classification tasks and to establish trends in the feature importance, we complement it with an explainable AI technique which provide insights into the range of the feature values that positively contribute spatter predictions. Furthermore, our ML model was applied and
tested against datasets generated from FLOW-3D, a prevalent yet computationally economical software extensively employed in AM.  However, in the standard FLOW-3D implementation used in this study, the surrounding gas is represented as a void region with prescribed thermodynamic conditions, rather than being explicitly resolved as a gas phase. This approximation reduces computational cost by avoiding direct solution of the Navier--Stokes equations for the surrounding gas, but it limits the ability to directly resolve gas--liquid interactions that influence spatter generation and transport \cite{myers2023high, FLOW-3D}. We note that tangential surface tension effects can be incorporated into FLOW-3D through custom user-defined implementation; therefore, the limitation discussed here refers specifically to the standard implementation used in this work. In contrast, the OpenFOAM framework employed in this study explicitly resolves key gas--liquid interactions and implements the tangential component of surface tension following the method by Wirth et al. \cite{wirth2022implementation}. By leveraging the respective strengths of OpenFOAM and FLOW-3D simulations, augmented with machine learning (ML), a reduced-order methodology was developed to accelerate spatter process map generation in LPBF (Figure \ref{fig:schematic}). This approach facilitates rapid and accessible quantification of generated spatter relative to the expected volume. This enables the creation of a comprehensive process map, providing insight into the spatter formation mechanism in additive manufacturing. This map can be used as a visualization tool to identify optimal process windows and parameters, enabling the printing of parts with low porosity and high-quality surface finish.

\FloatBarrier
\FloatBarrier

\section{Methods}

\subsection{Simulation of LPBF via OpenFOAM}

A CFD model was developed using OpenFOAMv2012 to analyze the complex physical phenomena that occur during the LPBF process. A newly introduced icoReactingMultiphaseInterFoam (IRMIF) solver \cite{wirth2022implementation, lundkvist2019cfd, msmsac4a26bib17} was used. The IRMIF solver is based on the volume-of-fluid (VOF) method, where each phase is immiscible, and a clear boundary between each phase is calculated. The IRMIF solver handles fluid mechanics, including turbulent flow, laser beam sources with arbitrary beam shapes, heat transfer, and phase transitions such as solidification, melting, and evaporation. However, the solver did not account for the Marangoni effect due to the absence of the tangential component of the surface tension. The presence of the tangential component could influence the shape of the melt pool, leading to denudation, spatter, and pore formations. The tangential component of the surface tension is added to the solver following the modification by Wirth et al. \cite{wirth2021influence} to incorporate the Brackbill equation \cite{msmsac4a26bib15} which enables the implementation of surface tension forces in the VOF method as shown in equation (1). The section that was added to the solver code (an excerpt of the phaseSystem.C file of the uncompiled solver source code) to account for the tangential surface tension is highlighted in the Figure \ref{fig:codeblock} of the appendix.
\begin{equation}
F_s = \left[ \sigma \kappa \vec{n} + \left( \nabla \sigma - \vec{n} (\vec{n} \cdot \nabla \sigma) \right) \right] \cdot \frac{2 \cdot \rho(x)}{\rho_1 + \rho_2} \cdot \frac{\left| \nabla \rho(x) \right|}{\rho_2 - \rho_1}
\end{equation}

 The parameters in Equation (1) include the surface tension $\sigma$, the interface curvature $\kappa$, the normal vector of the interface $\vec{n}$, and the densities $\rho_1$ and $\rho_2$ of the two phases at the interface. In brackets of Equation (1), the first term corresponds to the normal component of the surface tension force, while the second term corresponds to the tangential component. The factor $\frac{2 \cdot \rho(x)}{\rho_1 + \rho_2}$ acts as a weight function that redistributes the surface tension force smeared to the denser phase. The term $\frac{\left| \nabla \rho(x) \right|}{\rho_2 - \rho_1}$ determines the position of the interface where it is non-zero.

Using the VOF method, the dynamic geometry of the free surface (gas/ metal interface) was
captured. The volume fraction satisfies equation (2). In OpenFOAM, an additional convective term is introduced to the VOF, adapting it to a two-fluid Eulerian system \cite{cerne2001coupling}, which depends on the compression velocity $\vec{n}$. This velocity is active only in the interface cells due to the term $\alpha_1(1-\alpha_1$). This artificial velocity helps compress the interface by reducing the numerical diffusion of the volume fraction $\alpha_1$ while preserving its boundedness \cite{saldi2012marangoni}. Since only the VOF is solved by the solver used in this study, the forces at play drive the fluid flow, with the velocity determined through the momentum balance to account for ejection from the melt pool. Since the force acting on the particles is intrinsic to the fluid flow, as it is a viscous fluid rather than individual particles, solving the velocity field captures the dynamics of spatter ejection.
 \vspace{-2mm}
\begin{equation}
\internallinenumbers
\frac{\partial \alpha_1}{\partial t}=-\vec{\nabla} \cdot\left(\alpha_1 \vec{u}\right)-\vec{\nabla} \cdot \vec{u} \alpha_1\left(1-\alpha_1\right)
\end{equation}
In the situation of the mixture of vapor and nitrogen, the model applies a weighting average based on the composition of each component in the mixture in equation (3-7) to the value of its physical properties, such as density $\rho$, viscosity $\mu$, thermal conductivity $k$ and specific heat capacity $C$ as listed in Table 1. 
\begin{equation}
\alpha_1 + \alpha_2 = 1
\end{equation}
\begin{equation}
\rho = \alpha_1 \rho_1 + \alpha_2 \rho_2 \\
\end{equation}
\begin{equation}
\mu= \alpha_1 \mu_1 + \alpha_2 \mu_2 \\
\end{equation}
\begin{equation}
k = \alpha_1 k_1 + \alpha_2 k_2 \\
\end{equation}
\begin{equation}
C = \alpha_1 \frac{\rho_1}{\rho} C_1 + \alpha_2 \frac{\rho_2}{\rho} C_2
\end{equation}
The value of either $\alpha_1$ or $\alpha_2$ ranges from 0 to 1,  where $\alpha_1$ and $\alpha_2$ are the volume fractions of the phase of the element, $\vec{u}$ is the flow velocity, $t$ is the time. The finite volume method was used to compute the fluid flow, based on continuity and transient Navier-Stokes equation, as described in equations (8) and (9)  respectively.
 \vspace{-5mm}
\begin{equation}
\frac{\partial \rho}{\partial t} + \nabla \cdot (\rho \vec{u}) = 0
\end{equation}

 \vspace{-10mm}
 
\begin{equation}
\frac{\partial(\rho \vec{u})}{\partial t} + \nabla \cdot (\rho \vec{u} \otimes \vec{u}) = \nabla \cdot \left\{-p \cdot \vec{I} + \mu \left[\nabla \vec{u} + (\nabla \vec{u})^T\right]\right\} + \vec{F}
\end{equation}
where $\rho$ is the density, $\vec{u}$ is the fluid flow velocity vector, $t$ is the time, $p$ is the pressure, $I$ is the identity tensor, $\mu$  is the dynamic viscosity and $\vec{F}$ is the volume force vector. Rapid heating to high temperatures causes the molten material to evaporate, generating a recoil pressure within the melt pool. This recoil pressure, produced by the vapor jet, drives the melt surface downward, creating a vapor depression. The recoil pressure is modeled using the Clausius–Clapeyron equation to calculate the saturated vapor pressure, adjusted by a coefficient that accounts for the backward flux of the evaporated material, as shown in equation 10 \cite{klassen2014evaporation, ki2002modeling, heeling2017melt}.
\begin{equation}
\vec{P}_r = 0.54 P_0 \exp\left[\frac{L_v M (T - T_v)}{R T T_v}\right] \vec{n} \left|\nabla \alpha_1\right| \frac{2{\rho}}{\rho_{\text{metal}} + \rho_{\text{gas}}}
\end{equation}
where $P_0$ is the ambient pressure, $L_v$ is the latent heat of evaporation, $T_v$ is the evaporation temperature, $M$ is the molar mass, and $R$ is the universal gas constant. 
The energy equation in equation (11) was used to compute the temperature field.
\begin{equation}
\rho \frac{\partial E}{\partial t}=-\rho \vec{\nabla} \cdot \vec{u} E+\vec{\nabla} \cdot(k \vec{\nabla} T+\overline{\bar{\tau}} \cdot \vec{u}) + q_v + q_{\text {rad }}+q_{\text {laser }}
\end{equation}
where $E$ is the mixed energy, $k$ is the thermal conductivity, $T$ is the temperature, $\overline{\bar{\tau}}$ is the stress tensor. $q_\text{v}$ is the heat loss due to vaporization (Equation 12), $q_\text{rad}$ is the heat loss due to radiation (Equation 13), $q_\text{laser}$ is the heat input from the laser beam (equation 14). 
\begin{equation}
q_v=-0.82 \frac{l_v M}{\sqrt{2 \pi M R T}} P_0 \exp \left[\frac{l_v M\left(T-T_v\right)}{R T T_v}\right]\left|\nabla \alpha_1\right| \frac{2 \rho C_p}{\rho_{\text {metal }} \rho C_{\text {pmetal }}+\rho_{\text {gas }} C_{\text {pgas }}}
\end{equation}
\begin{equation}
    q_{\text {rad }}=-\sigma_s \varepsilon\left(T^4-T_{\text {ref }}^4\right)\left|\nabla \alpha_1\right| \frac{2 \rho C_p}{\rho_{\text {metal }} \rho C_{\text {pmetal }}+\rho_{\text {gas }} C_{\text {pgas }}}
\end{equation}
Where Where $C_p$ is the specific heat capacity, $L_m$ is the latent heat of melting, $\sigma_s$, $\varepsilon$, $T_{\text {ref }}$ are the Stefan-Boltzmann constant, emissivity and ambient temperature respectively.
The IRMIF solver uses the discrete transfer radiation model, which takes the laser beam into account, passing through VOF phases \cite{cao2020mesoscopic}. The laser energy was applied in the form of a body heat source and goes into the depth of the material. A Gaussian intensity distribution $(D_{4\sigma} = 100 \, \mu \text{m})$ is applied to the laser beam. 
\begin{equation}
q_{\text {laser }}=\frac{f_{\text {absorb }} I}{\Delta y} \quad \text { where } I=\frac{2 \xi P}{\Pi r^2} \exp \left[\frac{-2\left(x+v_0 t-x_0\right)^2+\left(z-z_0\right)^2}{r^2}\right]
\end{equation}

Where $I$ is the beam power density, $f_{\text {absorb }}$ is the percentage of laser energy occupied by the element, $P$ is the laser power, $\xi$ is the absorption coefficient, $r$ is the
beam radius, ($x_0$, $z_0$) describes the starting position of the laser beam, $v_0$ denotes the scanning
velocity and $\Delta y$ represents the size of the numerical grid.
IRMIF uses the Lee mass transfer model in equation (15) to calculate the solidification (or also melting) process, better described in \cite{bahreini2016development} and the original
formulation of Lee \cite{lee1980pressure}.
\begin{equation}
m=C^{\prime} \rho \alpha \frac{T-T_A}{T_A} L
\end{equation}
In equation 15,  $m$ is the mass transfer between phases, $C^{\prime}$ is Lee’s model constant, $\alpha$ is the phase volume fraction, $T$ is the current
temperature, $T_A$ is the solidification or melting temperature, $L$ is the latent heat. It should be noted
that the same set of equations works for both solidification ($T$ $<$ $T_A$ and $C^{\prime}$ negative) and melting
($T$ $>$ $T_A$ and $C^{\prime}$ positive) \cite{scuro2022progress}.
\begin{figure*}[htbp!]
\begin{center}
\includegraphics[width=1\linewidth]{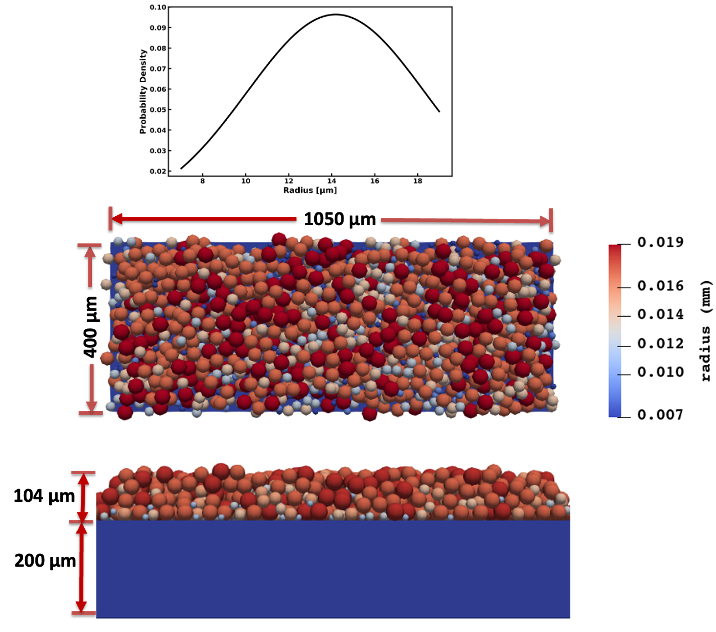}
\end{center}
    \caption{Schematic illustration of the computational domain with random packing of stainless steel powder.}
\label{fig:OpenFOAM_LPBF}
\end{figure*}

An open-source Discrete Element Method (DEM) code, LIGGGHTS (LAMMPS Improved for General Granular and Granular Heat Transfer Simulations)\cite{kloss2011liggghts}, was employed to simulate the random packing of SS316L particles deposited onto a solid substrate. The metal powders were modeled as solid spheres with a mean radius of 15 $\mu m$.
Following Newton’s second law of motion, the powder bed morphology was determined by computing the trajectories of spherical particles. Initially, a cloud of metal powders was generated just above a cubic domain. Under the influence of gravity and inter-particle contact forces, the particles freely settled into the domain, forming a densely packed powder bed. Particle interactions were governed by a Hertzian contact model~\cite{mindlin1953elastic} with tangential history and rolling friction. The simulation proceeded through four phases: particle insertion over \SI{1.5}{\second}, gravitational settling for \SI{1.0}{\second}, roller opening with combined linear and rotational motion over \SI{13}{\second}, and final particle discharge over \SI{1.0}{\second}. Simulation outputs were saved as VTK and STL files at intervals of \SI{0.05}{\second} for post-processing and visualization of particle kinematics and flow behavior. A thick layer of metal powders with a height of \SI{104}{\micro\meter} was ultimately generated.
  The positions and diameters of individual particles were extracted from the DEM simulation results and used to define the initial powder configuration in the subsequent Computational Fluid Dynamics (CFD) model as shown in  Figure \ref{fig:OpenFOAM_LPBF}. 
The dimension of the simulation domain is 1050 $\mu m$ $\times$ 400 $\mu m$ $\times$ 500 $\mu m$. Adaptive mesh was implemented in OpenFOAM where the mesh size along the x-axis and the y-axis is fixed at 2.5 $\mu m$ and 5 $\mu m$, respectively, while the z-axis varies from 2.5 $\mu m$ to 8.74 $\mu m$ to generate the total number of 2688000 cells.
\begin{table}[ht!]
\centering
\caption{Constant material properties implemented for the simulation of the SS316L melting process in OpenFOAM simulation \cite{heeling2017melt, bobkov2008thermophysical}, while the subscript $s$ represents the solid phase, $l$ for the liquid phase, $v$ for the vapor and $n$ for the shielding gas nitrogen} 
\begin{tabular}{@{}lllll@{}}
\toprule
Parameter                        & Value                                            & Units                   &  &  \\ \midrule
Density, $\rho$, 298 K              & 7618                                             & kg/m$^3$ &  &  \\
Density, $\rho$, 1923 K             & 6468                                            & kg/m$^3$ &  &  \\
Specific heat solid, $C_{p, s}$,         & 592.7                                              & J/kg/K                  &  &  \\
Specific heat liquid, $C_{p, l}$, 1923 K        & 1873                                              & J/kg/K                  &  &  \\

Specific heat vapor, $C_{p, v}$ & 775                                              & J/kg/K                  &  &  \\

Specific heat nitrogen, $C_{p, n}$ & 1035                                              & J/kg/K                  &  &  \\

Thermal conductivity solid, $k_s$, 298 K    & 24.57                                                & W/m/K                   &  &  \\

Heat of fusion, $Hf_{l}$    & 2.7$\times 10^5$                                             & J/kg                  &  &  \\

Heat of evaporation, $Hf_{v}$ & 449                                              & J/kg                  &  &  \\

Dynamic viscosity liquid, $\mu_{l}$                         & 3.16 $\times 10^3$                                         & J kg/m/s                  &  &  \\

Dynamic viscosity vapor, $\mu_{v}$                        & 1.79$\times 10^6$                                         & J kg/m/s                  &  &  \\

Prandtl number liquid, $\Pr_{l}$                        & 0.1214                                         &                   &  &  \\

Prandtl number vapor, $\Pr_{v}$                        & 0.7                                         &                   &  &  \\

Prandtl number nitrogen, $\Pr_{n}$                        & 0.7721                                         &                  &  &  \\
Surface Tension, $\sigma$          & 1.882 &          $kg/s^2$          &  &  \\
Liquidus Temperature, $T_L$       & 1723                                             & K                       &  &  \\
Solidus Temperature, $T_S$        & 1658                                             & K                       &  &  \\

Absorptivity        & 0.55                                             &                    &  &  \\

Latent Heat of Fusion, $\Delta H_f$            & 2.6 $\times 10^5$ & J/kg                    &  &  \\
Latent Heat of Vaporization, $\Delta H_v$    & 7.45 $\times 10^6$ & J/kg                    &  &  \\ \bottomrule
\end{tabular}
\label{table:ss316lmaterialparam}
 
\end{table}

It should be noted that the thermophysical properties used in this work correspond to SS316L. Therefore, the recoil pressure, surface tension force, Marangoni contribution, and temperature-dependent material response are implicitly material-specific. Since the present dataset is generated for a fixed material system, the resulting ML model should be interpreted as valid for the investigated SS316L conditions rather than as a universal model for all alloys. Extension of the framework to other materials would require material-specific simulations or the inclusion of additional material descriptors, such as surface tension, temperature coefficient of surface tension, absorptivity, viscosity, and evaporation parameters.

\subsection{FLOW-3D Simulation}
FLOW-3D (v11.2) simulations are performed to accelerate the development of the process map. FLOW-3D is a multiphysics simulation software produced by Flow Science, which provides more rapid estimations of the melt pool behavior than OpenFOAM. This reduction in computational expense is achieved in part by abstracting the dynamics of the vapor phase to estimates of the applied pressure and mass transfer at the surface of the melt pool. To create a FLOW-3D simulation data set, 281 SS316L single-track bare plate experiments are performed at varying processing parameters for a total length of 600 $\mu s$.

During simulation, the FLOW-3D package solves the equations that describe mass transfer, momentum transfer, and energy transfer during the melting process. This simulation is carried out on a structured Cartesian mesh, with mesh elements sized at 10 $\mu m$.  More specific information on the equations solved and the physical phenomena considered during the simulation can be found in \cite{ogoke2023convolutional, myers2023high, hemmasian2023surrogate, cook2020simulation, cheng2019computational}. In this framework, the OpenFOAM and FLOW-3D models are intentionally kept at a comparable level of particle--fluid treatment. OpenFOAM uses DEM (LIGGGHTS) to generate the powder bed, with particle--particle contact resolved during bed generation, and then solves the CFD on that fixed particle configuration; the FLOW-3D model
operates at a similar level. Neither solver performs fully coupled DEM--CFD during the melt-pool simulation, in which the fluid and the moving particles would exchange forces while the particles also interact with one another. This is a deliberate choice rather than an omission: because the machine learning model is trained on OpenFOAM data and applied to FLOW-3D outputs, the two datasets must remain physically comparable for the transfer to remain valid. Introducing fully coupled DEM--CFD into OpenFOAM alone would increase the discrepancy between the two solvers and degrade the cross-platform transfer that underpins the process map. Moreover, fully coupled DEM--CFD is most relevant to powder-entrainment spatter, which is outside the present scope, rather than to the recoil-driven hot spatter studied here. It is therefore identified as future work that would also require a corresponding reduced-order treatment on the FLOW-3D side.

\subsection{Spatter Extraction and Analysis}
\begin{figure*}[htbp!]
\begin{center}
\includegraphics[width=1\linewidth]{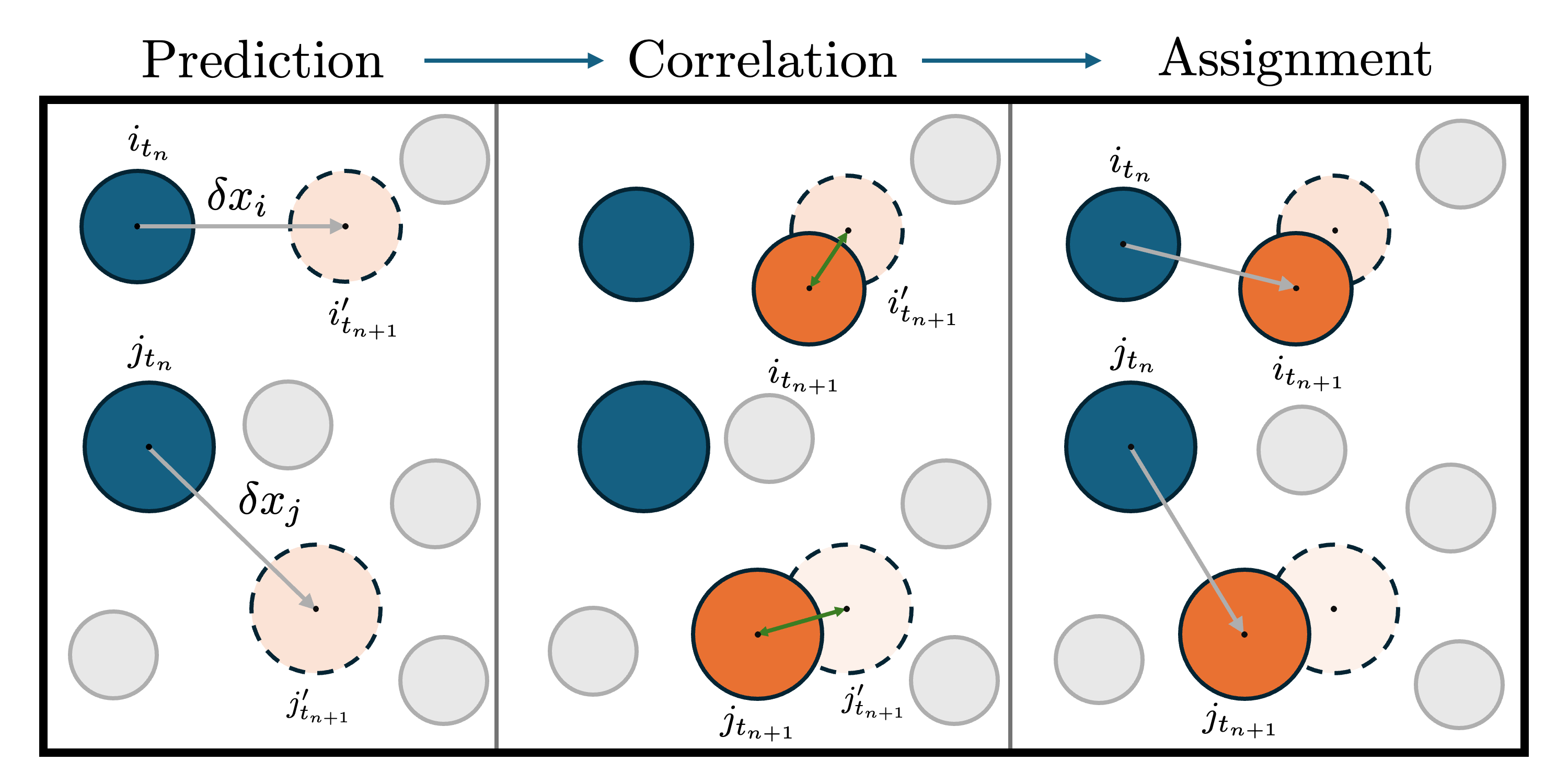}
\end{center}
    \caption{A tracking algorithm is implemented to consistently identify  spatter particles as they move through the domain. The tracking process is demonstrated for two sample particles $i$ and $j$, between the timesteps ${n}$ and ${n+1}$. During the tracking process, the velocity field of the particle is used to predict an estimated displacement $\delta x_{i}$ within $t_{n+1} - t_{n} = \Delta t$. Next, the actual particle positions at the next timestep, $i_{t_{n+1}}$,  are correlated to the estimated displacement values $i'_{t_{n+1}}$ based on a nearest neighbors assignment. Following this, each particle $i_{t_n}$ is linked to $i_{t_{n+1}}$, completing the tracking process.}
\label{fig:tracking}
\end{figure*}

The OpenFOAM simulation data are processed to extract properties of both the ejected spatter and the melt pool. The cell-level simulation information is extracted for processing in Python, where each field is represented as a three-dimensional matrix defined by the mesh. As part of this information, three fields -- $\alpha_g$, $\alpha_s$ and $\alpha_l$ denote the volume fraction of the gas phase, the solid phase, and the liquid phase, respectively. To differentiate the powder bed, spatter, and melt pool from the gaseous phases of the simulation, the gas volume fraction $\alpha_g$ is truncated at 0.5, masking cells in the domain occupied by inert gas or metal vapor from the current analysis.

In this scenario, the ejected fluid spatter droplets as particles in our studies to denote a connected set of fluid cells that are disconnected from the larger melt pool. Connected component analysis \cite{fiorio1996two} is used to isolate individual spatter particles for further analysis. To do so, a binary field is constructed from $\alpha_{g}$ by applying a threshold at 0.5, assigning 1 to solid cells and liquid cells and 0 to gas phase cells. The spatter particles, powder bed and melt pool are then identified by applying the Union-Find algorithm to the binary array, which assigns unique labels to each disconnected object within the 3D domain \cite{galler1964improved, scikit-image}. The largest contiguous object is identified as the composite melt pool and powder bed structure, and all other objects above a size-based noise threshold are considered to be spatter particles. Specifically, this threshold acts as a lower bound on the size of the particles, where particles comprised of less than eight connected cells are removed from the analysis. Following this processing step, the mesh elements that construct each particle can be used to extract particle-level field information.

To extract the kinematics of the particle behavior, the positions of the particles must be tracked across the evolution of the simulation. During the connected components analysis, individual particles are identified without any intrinsic order. Therefore, an additional processing step is required to identify the mapping between particles at subsequent timesteps. We perform this tracking operation as a three-stage process -- prediction, correlation and assignment. Given a set of $m$ particles at timestep $t_n$ and a set of $k$ particles at timestep $t_{n+1}$, the goal of this process is to associate each of the $m$ particles at $t_n$ with either the index of the corresponding particle at $t_{n+1}$, or a condition indicating that the particle trajectory has ended. The remaining particles at timestep $t_{n+1}$ that have not been associated with a particle at timestep $t_{n}$ are considered to have formed at timestep $t_{n+1}$ (Figure \ref{fig:tracking}).

During the first phase of the tracking process, the velocity of each particle is used to estimate the displacement that occurs over a time $\Delta t$. Specifically, the average of the velocity fields for the cells comprising each particle is used to estimate the displacement $\delta \mathbf{x}_i$ (Equation \ref{equation_averaging}), assuming a constant velocity over the timestep $\Delta t$. In Equation \ref{equation_averaging}, $N_s$ refers to the number of cells comprising a particle, and $\mathbf{v_s}$ refers to the velocity of each individual cell.

\begin{equation}
\delta \mathbf{x}_{i} = \frac{\Delta t}{N_{s}} \sum_{s = 1}^{N_{s}} \mathbf{v_s} 
\label{equation_averaging}
\end{equation}

This displacement defines an estimate of the position of each particle at time $t_{n}$ at the next timestep,
\begin{equation}
\mathbf{x}_{i'}|_{t_{n+1}} = \mathbf{x}_{i}|_{t_{n}} + \delta \mathbf{x}_i
\end{equation}

For each of these estimates $\mathbf{x_{i'}}$, we find the nearest neighbor to the estimate within the set of $k$ particles at timestep $t_{n+1}$. If this nearest neighbor lies within a specified maximum distance $\Delta x$, the nearest neighbor is assigned to have the same index $i$. However, if no such neighbor is found, the trajectory is assumed to have ended with the particle either leaving the domain or impacting the powder bed. To improve the computational efficiency of the tracking process, a $k-d$ tree structure is queried to identify the pairwise distances between the sets of particles considered. 
With the tracking process complete, the kinematic properties of each particle, such as the magnitude and direction of motion, can be computed.

To sample points from the melt pool surface, the volume fractions of the gas and liquid phases are used to define the melt pool morphology. Specifically, $\alpha_{g}$ is first used to threshold and identify the partially molten powder bed. From the points comprising the partially molten powder bed, the melt pool is identified by isolating values with a liquid volume fraction $\alpha_{l}$ greater than 0.5. The surface is then identified by finding the grid cells with the highest value of $\alpha_g$ within the points comprising the melt pool. During the sampling process, we extract the average of a specific field within a $n_r \times n_r \times n_r$ bounding cube centered at a specific point along the surface, only taking into account the points within the bounding cube that lie within the melt pool, where $n_r$ is an arbitrarily selected parameter describing the size of the region investigated. We sample $n_{i}$ points from the melt pool at each timestep, where $n_{i}$ is the number of spatter particles ejected at time $i$. Each sample contains the mean of the position, velocity, density and temperature of the sampled region. Once this data has been sampled, the properties of the sampled regions are compared to the properties of the spatter particles.

\subsection{Machine Learning Methodology}
\subsubsection{\normalfont Data Collection and Hyper-Parameter Optimization}
Our dataset was extracted from OpenFOAM simulations of three experiments with the following power/velocity combinations: $545~\mathrm{W} / 2~\mathrm{ms}^{-1}$, $600~\mathrm{W} / 1~\mathrm{ms}^{-1}$, and $690~\mathrm{W} / 2~\mathrm{ms}^{-1}$.

 The spatter and the melt pool features were extracted using ParaView. The extracted features were used to validate our developed tracking algorithm, which is efficient in extracting not only spatter and melt pool properties but also the capability to track spatter trajectory. Detecting variations in the property distributions between the ejected spatter and the melt pool can enhance classification tasks. We implement different ML models as discussed in the Appendix A. to establish the performance trends for the spatter/melt pool classification task. From OpenFOAM experiments, we collected 323 data points, evenly split between the two classes (50 \% spatter and 50 \% melt pool). For the ML tasks, the dataset was divided into training (80 \%) and testing (20 \%) subsets using train test split, with a fixed random seed to ensure reproducibility of the results. Hyperparameter optimization was performed using GridSearchCV, using the ROC-AUC score as the evaluation metric. The grid search explored a predefined hyperparameter space as listed on Table 2, the best parameters were identified by selecting the combination that maximized the average ROC-AUC score. Predictions and probabilities were generated for both the training and testing sets. Model performance was evaluated using metrics including accuracy, F1-score, balanced accuracy, and ROC-AUC to assess the model’s predictive performance and generalization ability.

\begin{table}[h!] % Use table environment
\centering
\begingroup
\setlength{\tabcolsep}{4pt} % Reduce column spacing
\renewcommand{\arraystretch}{1.0} % Adjust row spacing
\setstretch{0.5} % Set line spacing to 0.5
\caption{Grid search hyperparameters range and the best value used for ML models.}
\begin{tabular}{@{}p{2.5cm}p{7.5cm}p{5.5cm}@{}}

\toprule
\textbf{Model} & \textbf{Grid Search Parameters} & \textbf{Used/Best Parameters} \\ 
\midrule

\textbf{RF} & 
\texttt{\{ 'n\_estimators': range(1, 100), 'max\_depth': range(3, 12), 'max\_features': ['sqrt'],  'criterion': ['gini', 'entropy'] \}} & 
\texttt{\{ 'criterion': 'entropy', 'max\_depth': 7, 'max\_features': 'sqrt', 'n\_estimators': 20 \}} \\ 

\midrule
\textbf{XGB} & 
\texttt{\{ 'n\_estimators': [50, 100], 'learning\_rate': [0.01, 0.1], 'max\_depth': range(3, 5), 'subsample': [0.8, 1.0], 'colsample\_bytree': [0.8, 1.0], 'gamma': [0, 1], 'reg\_lambda': [1, 10] \}} & 
\texttt{\{ 'learning\_rate': 0.1, 'max\_depth': 5, 'colsample\_bytree': 1.0,  'gamma': 1.0,  'reg\_lambda': 1, 'subsample': 0.8,  'n\_estimators': 150 \}} \\ 

\midrule
\textbf{Bagging} & 
\texttt{\{ 'n\_estimators': [5, 10], 'max\_samples': [0.7, 0.8], max\_features': [0.7, 0.8], 'bootstrap': [True, False], 'base\_max\_depth': [10, 20] \}} & 
\texttt{BaggingClassifier( base\_estimator=extra\_tree, 'base\_max\_depth': 20, 'bootstrap': True, 'max\_features': 0.8, 'max\_samples': 0.8, 'n\_estimators': 5} \\ 

\midrule
\textbf{LGBM} & 
\texttt{\{ 'n\_estimators': [100, 200], 'learning\_rate': [0.01, 0.1], 'max\_depth': [10, 20], 'num\_leaves': [31, 50], 'min\_child\_samples': [10, 20], 'subsample': [0.8], 'colsample\_bytree': [0.8] \}} & 
\texttt{\{ 'colsample\_bytree': 0.8, 'learning\_rate': 0.1, 'max\_depth': 20, 'min\_child\_samples': 10, 'n\_estimators': 200, 'num\_leaves': 31, 'subsample': 0.8 \}} \\ 
  \midrule\textbf{KNN} & 
\texttt{\{ 'n\_neighbors': range(3, 31, 2), 'weights': ['uniform', 'distance'], 'p': [1, 2], 'metric': ['minkowski'] \}} & 
\texttt{\{ 'n\_neighbors': 5, 'p': 1, 'weights': 'distance', 'metric': 'minkowski' \}} \\ 

\midrule
\textbf{SVC} & 
\texttt{\{ 'C': [0.1, 1, 10], 'kernel': ['linear', 'rbf', 'poly'], 'gamma': ['scale', 'auto'], 'degree': [2, 3] \}} & 
\texttt{\{ 'C': 10, 'degree': 3, 'gamma': 'scale', 'kernel': 'rbf' \}} \\ 

\bottomrule
\end{tabular}
\label{table:hyperparameter}
\endgroup
\end{table}

 \subsubsection{\normalfont SHapley Additive exPlanations as an Explainable AI}

SHapley Additive exPlanations (SHAP) implement a game-theoretic approach to describe how the ML models make predictions. It is based on the concept of Shapley values from cooperative game theory, where contributions of individual players (features) to the overall game outcome (model prediction) are calculated fairly \cite{lundberg2017unified}. SHAP values assign each feature a contribution value for every prediction, ensuring both global and local interpretability. The method guarantees local accuracy (the sum of SHAP values equals the model prediction) and consistency (a feature with increased contribution in a newer model will not have a lower SHAP value). 

Mathematically, the SHAP value for a feature $i$ is given by:
\begin{equation}
\phi_i = \sum_{S \subseteq N \setminus \{i\}} \frac{|S|! (|N| - |S| - 1)!}{|N|!} \big[f(S \cup \{i\}) - f(S)\big],
\label{SHAP_equation}
\end{equation}

where $S$ is a subset of features, $N$ is the set of all features, and $f(S)$ represents the model prediction with features in $S$. This ensures that SHAP fairly distributes the prediction among features by evaluating all possible subsets.

SHAP provides two key benefits: (1) Global interpretability, where aggregated SHAP values reveal feature importance across the dataset, and (2) Local interpretability, where SHAP values explain individual predictions by quantifying each feature's contribution.

 Partial Dependence Plots (PDPs) provide a visual representation of how one or two features influence model predictions, independent of other features. For a single feature $X_j$, the partial dependence function is defined as \cite{geurts2006extremely}:

\begin{equation}
\text{PDP}(X_j) = \frac{1}{n} \sum_{i=1}^n f(X_{i, -j}, X_j)
\label{PDP_equation}
\end{equation}
where $X_{i, -j}$ denotes all features except $X_j$. Single-feature PDPs illustrate the marginal effect of a feature, while pairwise PDPs show the joint effect of two features on predictions. PDPs are particularly useful for validating model interpretability and identifying nonlinear relationships.

\section{Results and Discussion}
\subsection{Validation of Numerical Simulations}
Mesh independence experiments were carried out on OpenFOAM and FLOW-3D as shown on the Figure \ref{fig:mesh_study} of the Appendix.  Mesh sizes of  5 $\mu m$ and 10 $\mu m$ were compared in FLOW-3D, while mesh sizes of 2.5 $\times$ 5 $\times$ 2.5-8.74 $\mu m$,  5 $\mu m$ and 10 $\mu m$ were compared in OpenFOAM. Close matches were observed in OpenFOAM and FLOW-3D, indicating that the mesh resolves heat transfer at the chosen resolution across the mesh sizes investigated.

The simulation parameters and physics in the FLOW-3D and OpenFOAM models have been calibrated to match the experimental thermal results, following the methodology proposed by Myers and Quirarte et al. \cite{myers2023high}. In these experiments, SS316L single-bead tracks were printed without powder using a TRUMPF TruPrint 3000 L-PBF machine. High-speed imaging of the melt pool was achieved using a Photron FASTCAM mini-AX200 color camera, capturing images at a rate of 22,500 frames per second, where each pixel corresponds to an area of 5.6 $\mu m$. These images were subsequently converted into pixel-by-pixel temperature data using the two-color method, which helps to reduce the impact of emissivity's temperature dependence by using the intensity ratios from the camera's red and green channels. Myers and Quirarte et al. \cite{myers2023high} provide detailed experimental methodologies and results. The dimensions of the melt pool are determined by sectioning, etching, and measuring the dimensions of the single-bead tracks.

These comparisons are conducted at the laser powers that are within the range at which our OpenFOAM simulations for these studies were carried out (150 W, 300 W, and 450 W), with the scanning speed maintained at $1~\mathrm{ms}^{-1}$ as shown in Table 1. At 150 W and 450 W while the scanning speed is 1 m/s, the width and depth of OpenFOAM simulation and experiment results are within the same range, while at 350 W and scanning speed of 1 m/s, there were slight variations. On comparing OpenFOAM melt pool dimensions to the lower bound of the width and depth of the experimental measurements at 300 W and $1~\mathrm{ms}^{-1}$,  deviations of 7.8 \% and 1.3 \% respectively were observed.  These results show close agreements between experimental studies and OpenFOAM with respect to the measured depth and width. The deviation may be due to the experimental settings of the laser orientation as it was positioned at an angle of approximately 20$^{\circ}$ during the interaction with the build plate for the accommodation of the imaging apparatus \cite{myers2023high}. Ultimately, developing a process window using the Flow3D dataset via the ML model trained on the OpenFOAM dataset requires measurement agreements between both simulations. Slight variations with respect to the width and depth are observed with the highest being the measured depth at 450 W with 7.8 \% decrease from OpenFOAM to Flow3D simulations.
\begin{table}[h!]
\caption{The melt pool dimensions comparison of experiment, FLOW-3D and OpenFOAM at a fixed speed of $1~\mathrm{ms}^{-1}$.The experimental widths and depths provided are generated from the solidified melt track, while the reported errors are in two standard deviations.}

\centering
\begin{tabular}{@{}llll@{}}

\toprule
         Power [W]            & 150  & 300 & 450 \\ \midrule
Experimental width ($\mu m$)  &   105 ± 7.98   &  161 ± 9.00 & 181 ± 4.62       \\
Flow3D width ($\mu m$)  & 95 & 138  & 180 \\         
OpenFOAM width ($\mu m$) &   100    &  140      &   176 \\
Experimental depth ($\mu m$)   &  31.0 ±  3.32 & 78.9 ± 4.86  & 155 ± 8.58       \\
Flow3D depth ($\mu m$) & 35 & 80 & 140            \\
OpenFOAM depth ($\mu m$) &   38    &   73     &    152 \\\bottomrule
\label{table:validation}
\end{tabular}

\end{table}
Further validation via comparison between the FLOW-3D, OpenFOAM data and experimental temperature profiles is displayed in Figure 3 for the set of process conditions, where the experimental measurements combine images captured at different camera exposure times (1.05 $\mu$s, 1.99 $\mu$s, 6.67 $\mu$s, and 20 $\mu$s). Experimental temperature values below 2000 $^{\circ}$C are not visible via high-speed imaging could be due to the low intensity of the green and red thermal radiation at low temperatures. The deviation in the experimental measurements is due to the plume emission and transmission above the melt pool that interferes with temperature measurement and obscure the front and side melt pool boundaries with temperature. However,  a close range and similar rise-and-fall profile can be observed between the experimental and simulation results, particularly for the larger section where the experimental melt pool dataset was collected \cite{myers2023high}.
In addition to validating the dimensions and temperature distribution of the melting pool, the dynamics of the spatter simulated in OpenFOAM was assessed against the diameter and velocity experimentally reported by Ly et al.~\cite{ly2017metal}.  The simulated recoil spatter across a range of ejection particle diameters and velocities  falls within the experimentally observed ranges as shown in Figure~\ref{fig:SonnyLy}a and Figure~\ref{fig:SonnyLy}b, respectively. Furthermore, the simulation yields an overall mean diameter of \(39.6~\mu\mathrm{m}\) and a mean ejection velocity of  \(5.5~\mathrm{m\,s^{-1}}\) while the experimental data reported by Ly et al. show an overall mean diameter of approximately \(35.1~\mu\mathrm{m}\) and an overall mean velocity of \(5.1~\mathrm{m\,s^{-1}}\). The close agreement between the simulated and experimental values demonstrates that the OpenFOAM model accurately reproduces the characteristic dynamics of recoil spatter.

\begin{figure*}[h!]
\begin{center}
\includegraphics[width=1\linewidth]{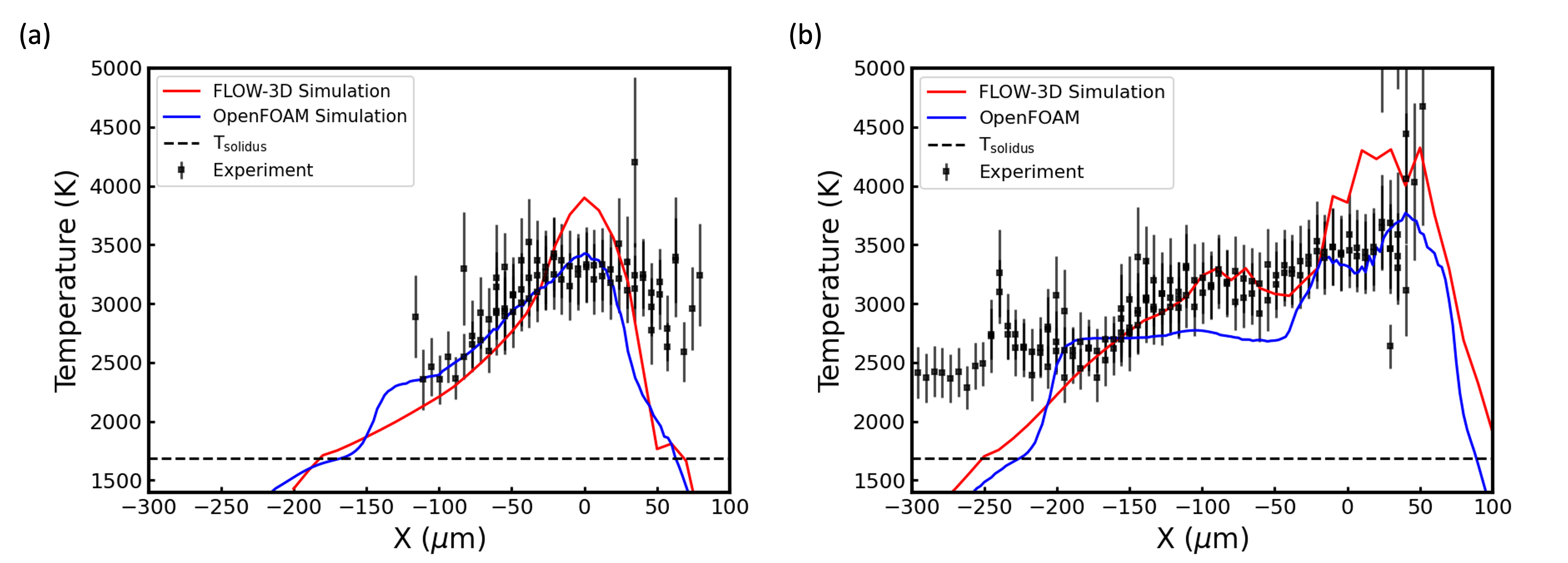}
\end{center}
    \caption{Implementation of the two-color thermal imaging method as described in \cite{myers2023high} for the comparison of the observed surface temperatures for SS316L with the corresponding surface temperatures obtained from modeling packages: a) Measurements taken at P = 150 W, V = $1~\mathrm{ms}^{-1}$ b)Measurements taken at P = 300 W, V = $1~\mathrm{ms}^{-1}$. The error bars from the experimental measurements represent combined uncertainties in emissivity and signal, as detailed in \cite{myers2023high}. }
\label{fig:validation}
\end{figure*}
\begin{figure*}[h!]
\begin{center}
\includegraphics[width=1\linewidth]{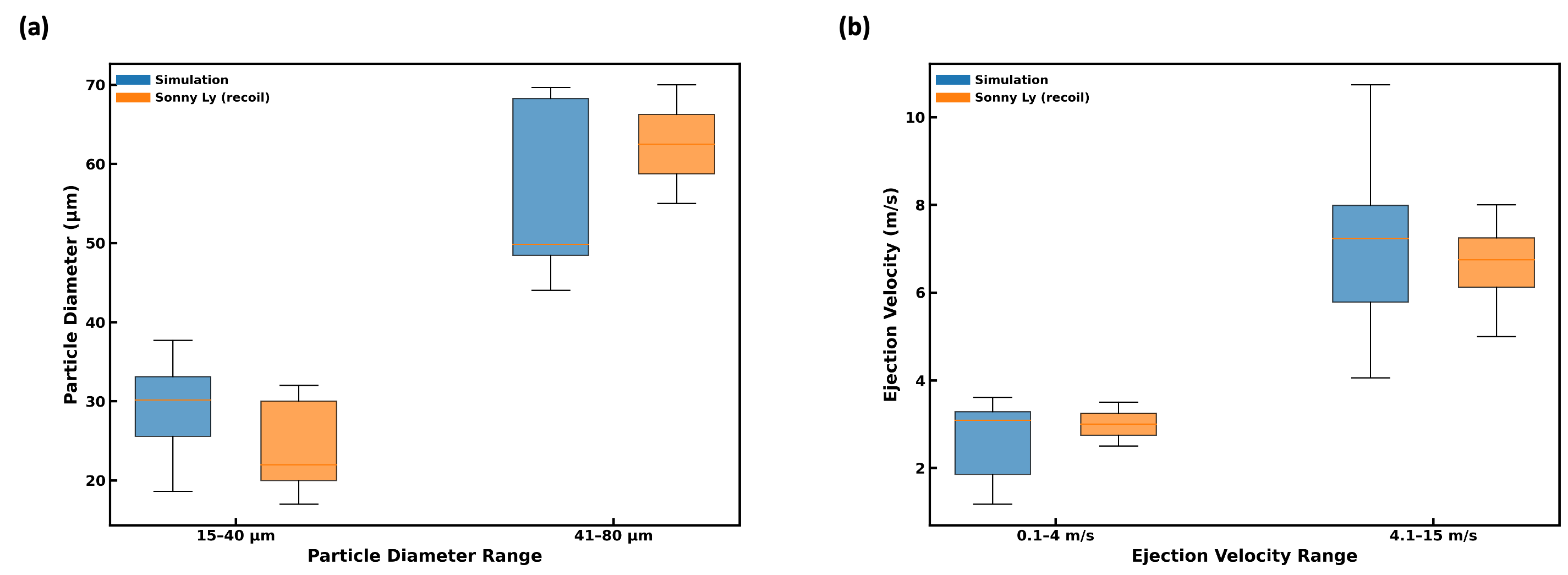}
\end{center}
    \caption{Comparison of simulated and experimental spatter dynamics reported by Ly et al. \cite{ly2017metal}. (a) Particle diameter. (b) Ejection velocity.} 
\label{fig:SonnyLy}
\end{figure*}
\begin{figure*}[htbp!]
\begin{center}
\includegraphics[width=0.8\linewidth]{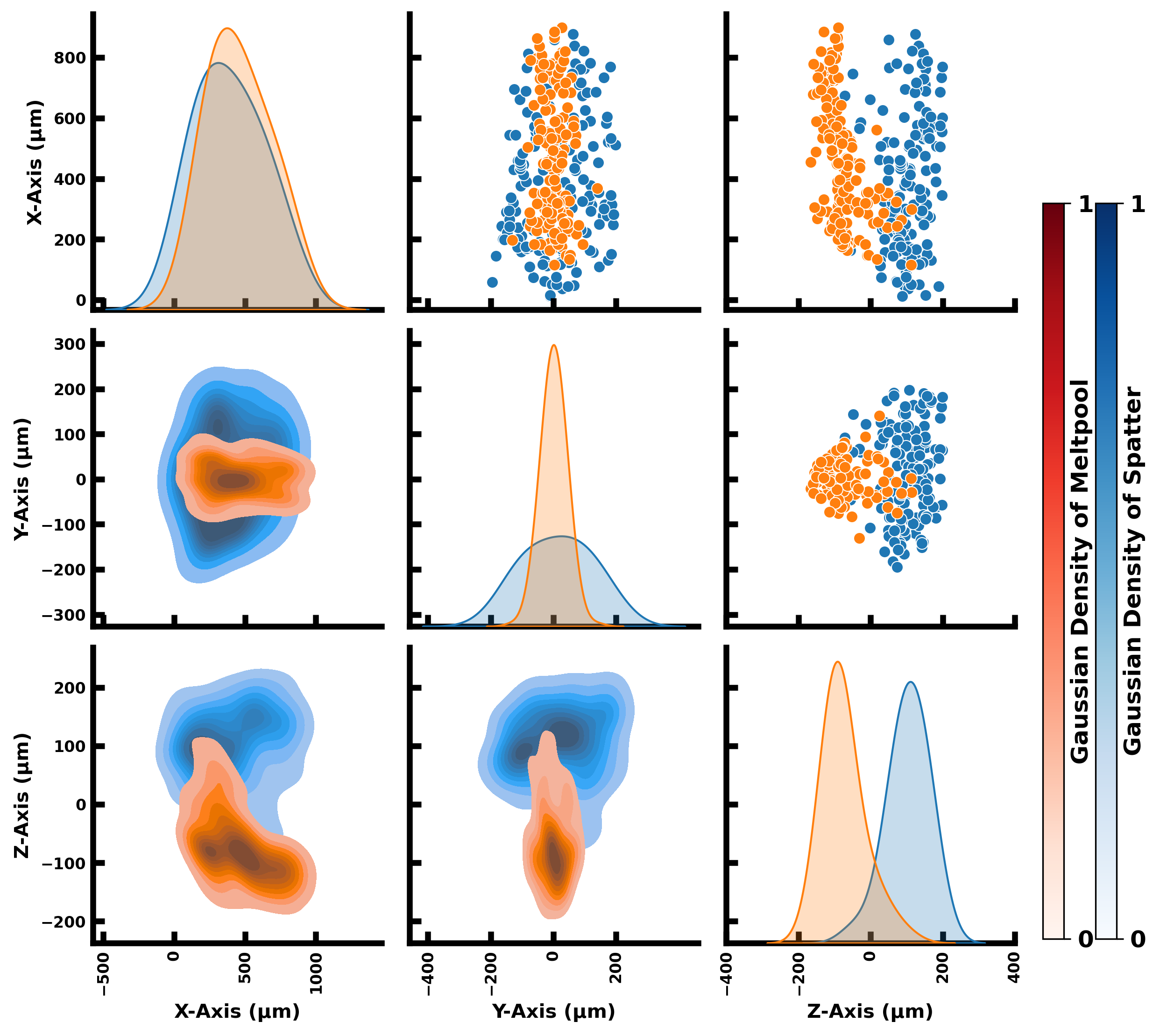}
\end{center}
    \caption{Pairplot correlating spatial distribution of the melt pool and the spatter ejection through the histogram along the diagonal, Kernel Density Estimation (KDE) plots to the left of the diagonal and scatter plot to the right of the diagonal . }
\label{fig:pairplot_coordinate}
\end{figure*}

\begin{figure*}[htbp!]
\begin{center}
\includegraphics[width=1.1\linewidth]{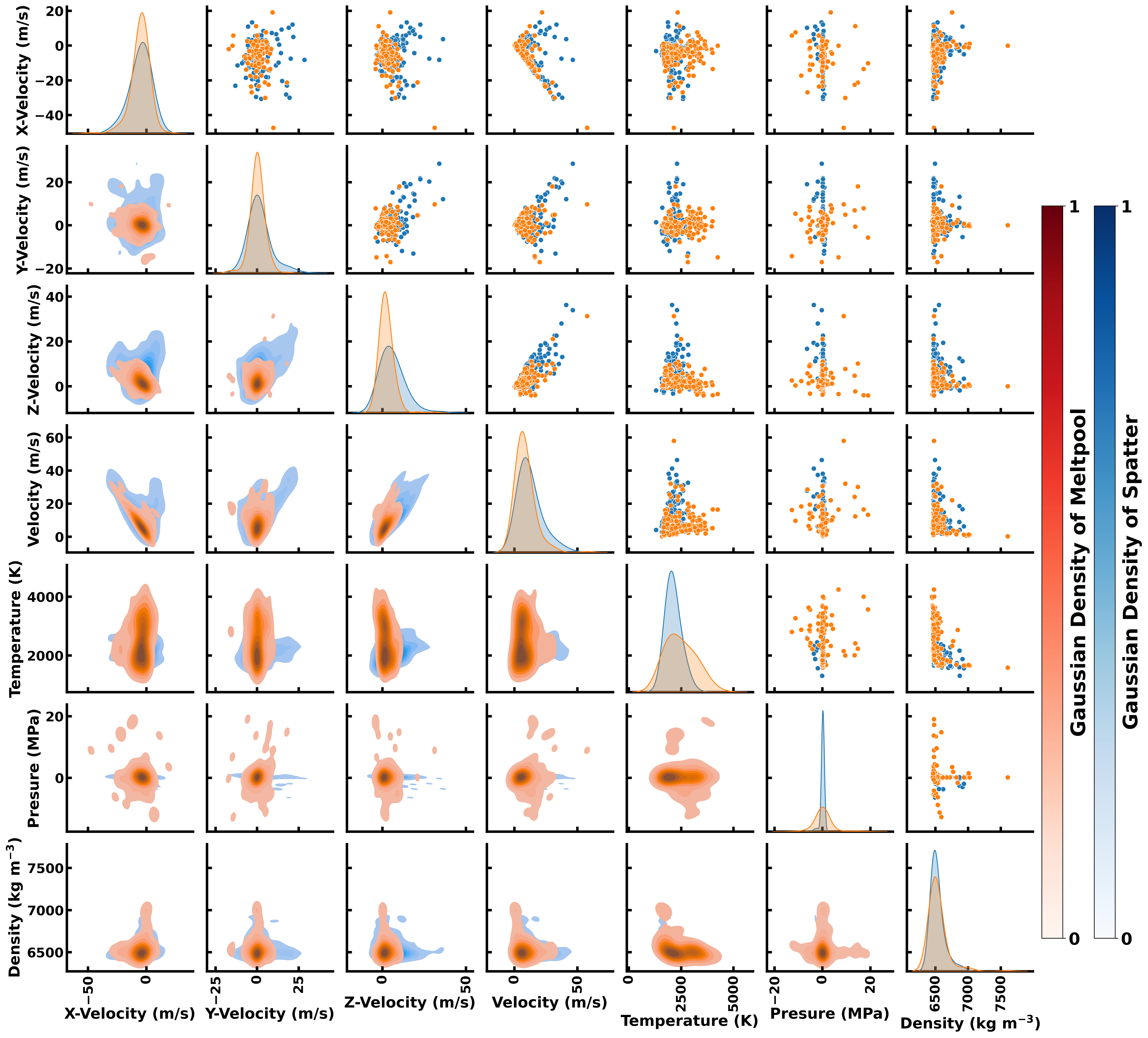}
\end{center}
    \caption{Pairplot correlating different measured properties of the melt pool and the spatter ejection through the histogram along the diagonal, Kernel Density Estimation (KDE) plots to the left of the diagonal and scatter plot to the right of the diagonal.  }
\label{fig:pairplot}
\end{figure*}

\begin{figure*}[htbp!]
\begin{center}
\includegraphics[width=1.1\linewidth]{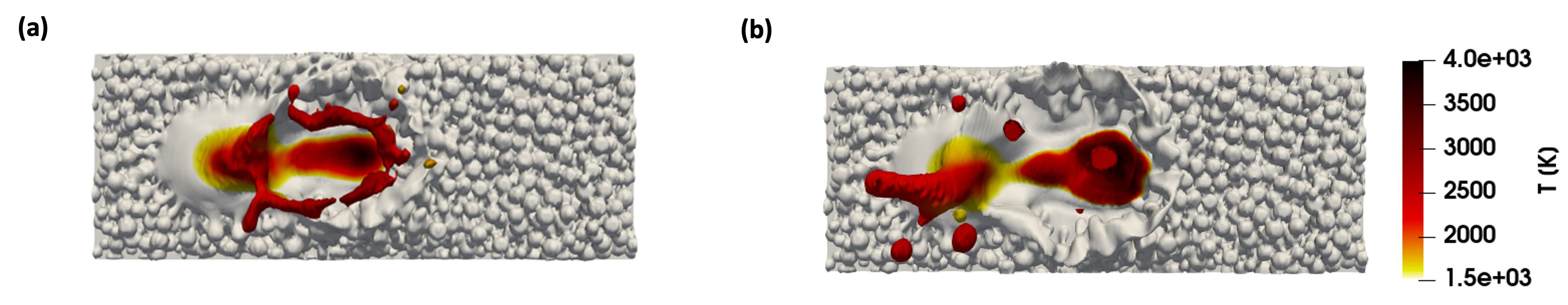}
\end{center}
    \caption{ OpenFOAM simulation at $545~\mathrm{W} / 2~\mathrm{ms}^{-1}$ displaying spatter event at different timestep (a) spatter event 170 $\mu s$ (b) Spatter event at 200 $\mu s$. }
\label{fig:Openfoam}
\end{figure*}

\begin{figure*}[htbp!]
\begin{center}
\includegraphics[width=1\linewidth]{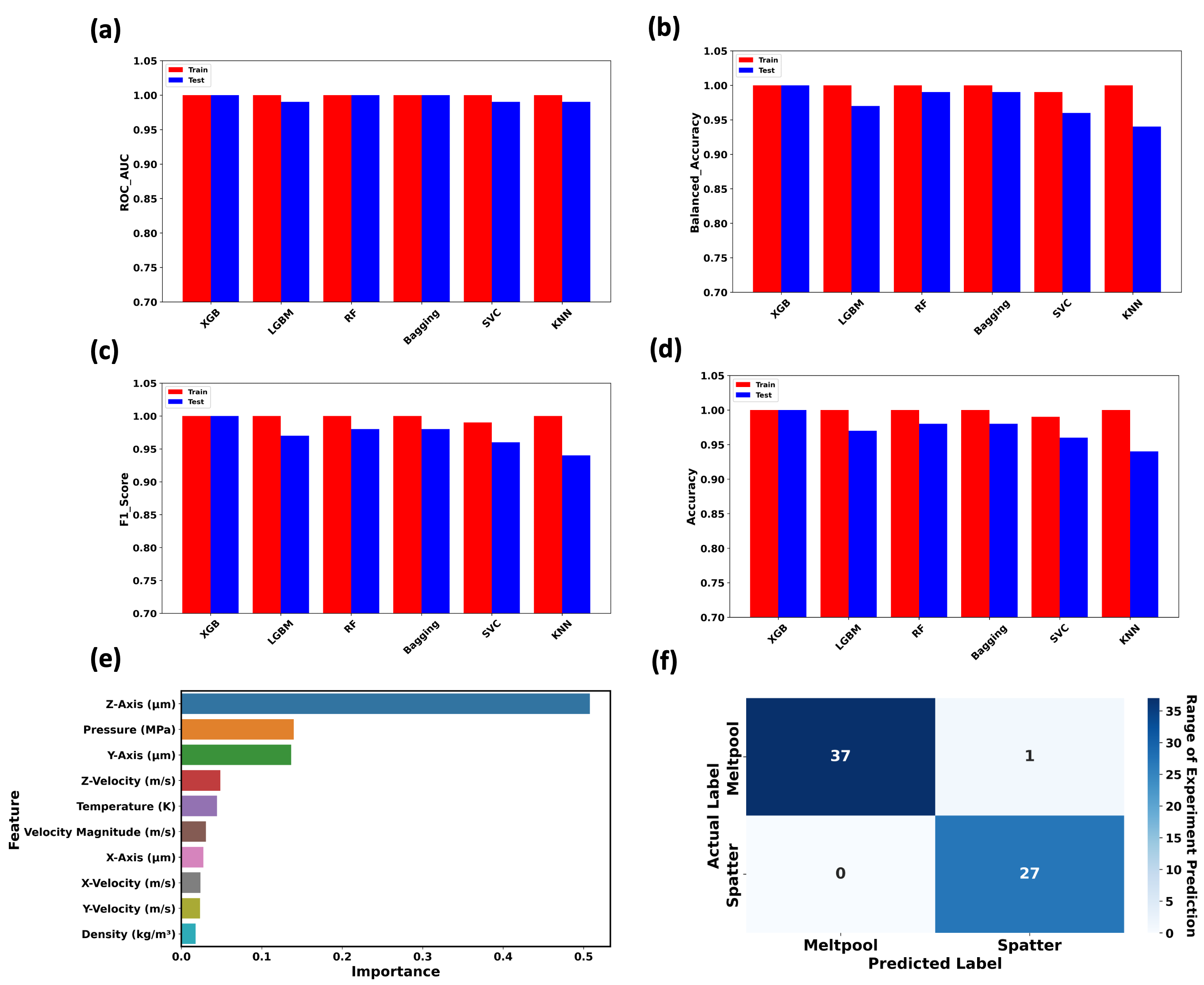}
\end{center}
    \caption{Performance measurement of ML models on classifying spatter ejection and melt pool with the following features: x-axis, y-axis, z-axis, temperature, velocity, x-velocity, y-velocity, z-velocity, density and pressure (a) ROC\_AUC score (b) Balanced\_Accuracy score (c) F1 score (d) Accuracy (e) Feature importance measurement from RF model (f)  Quantification of the correct and wrong predictions of spatter and melt pool from RF model.}
\label{fig:Corrdinate_Accuracy}
\end{figure*}

\begin{figure*}[htbp!]
\begin{center}
\includegraphics[width=1\linewidth]{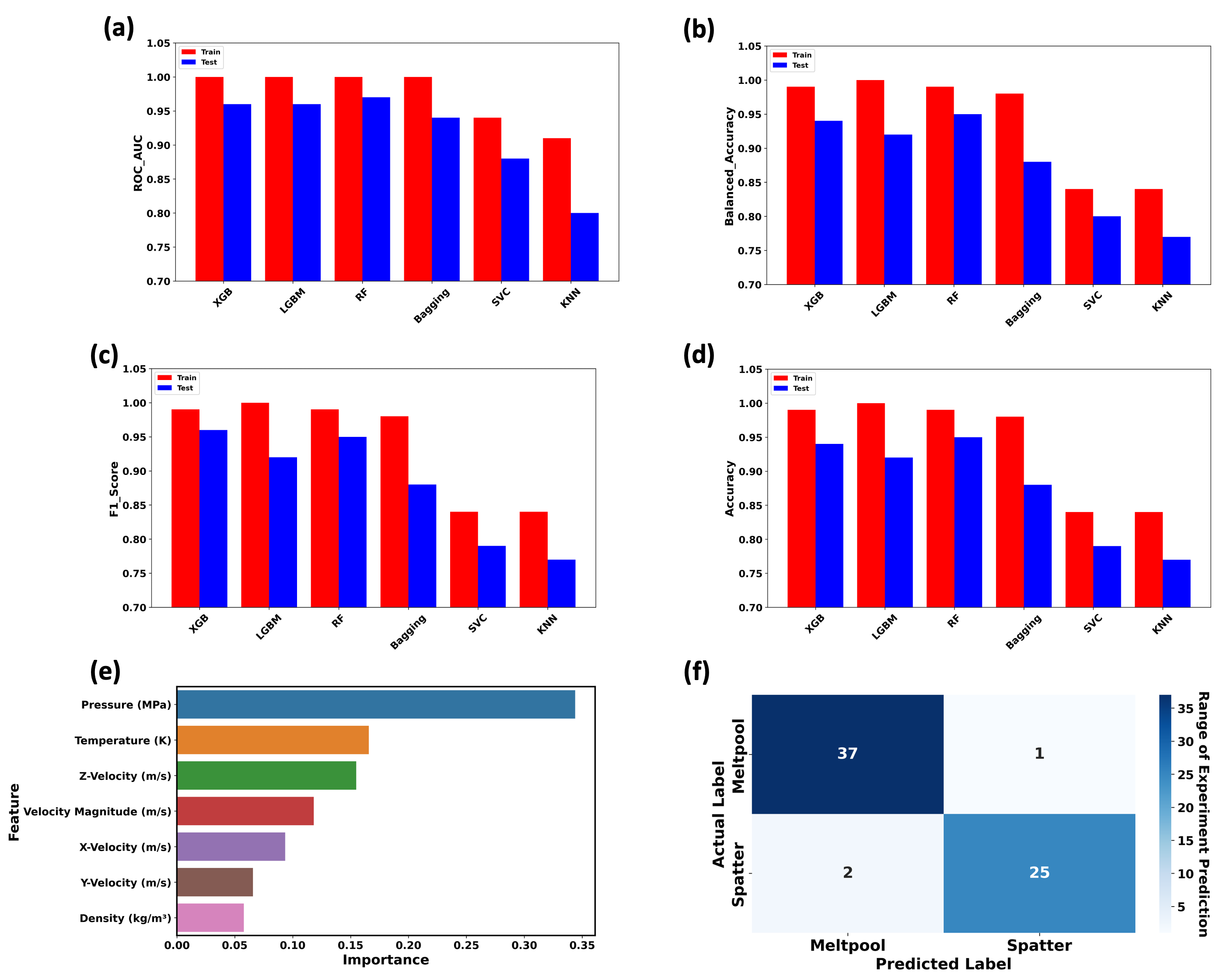}
\end{center}
    \caption{Performance measurement of ML models on classifying spatter ejection and melt pool with the following features:  temperature, velocity, x-velocity, y-velocity, z-velocity, density and pressure (a) ROC\_AUC score (b) Balanced\_Accuracy score (c) F1 score(d) Accuracy (e) Feature importance measurement from RF model (f)  Quantification of the correct and wrong predictions of spatter and melt pool from RF model.}
\label{fig:Accuracy}
\end{figure*}

\begin{figure*}[htbp!]
\begin{center}
\includegraphics[width=1\linewidth]{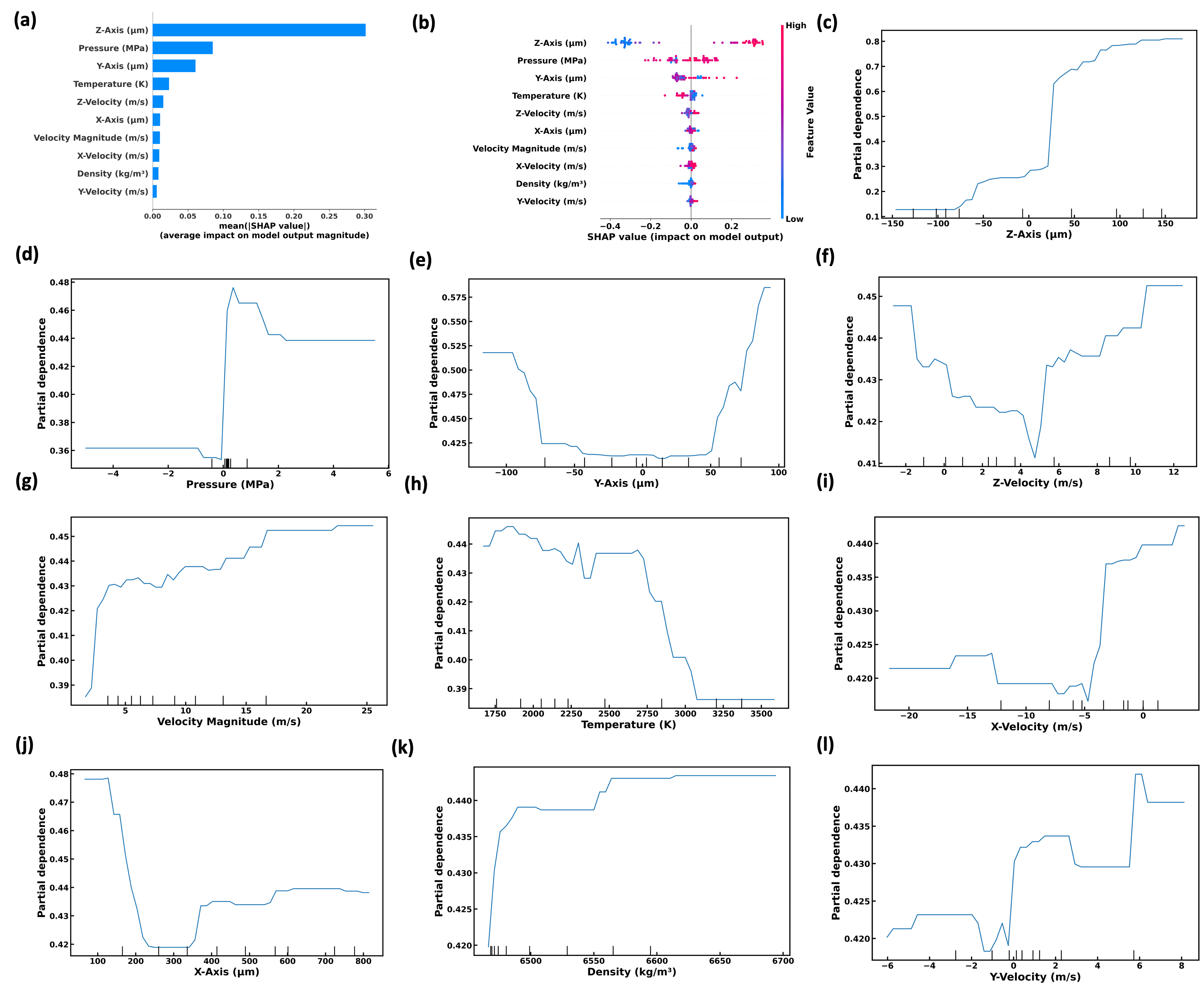}
\end{center}
    \caption{SHAP analysis on RF model in classifying spatter ejection and melt pool with the following features:  temperature, velocity, x-velocity, y-velocity, z-velocity, density and pressure (a) SHAP feature important (b) The SHAP summary plot with a color bar (c)  pressure PDP (d) z-velocity PDP (e) temperature PDP (f)   x-velocity PDP (g) Velocity magnitude PDP (h) Density PDP (i) y-velocity PDP}
\label{fig:Shap_coordinate}
\end{figure*}

\begin{figure*}[htbp!]
\begin{center}
\includegraphics[width=1\linewidth]{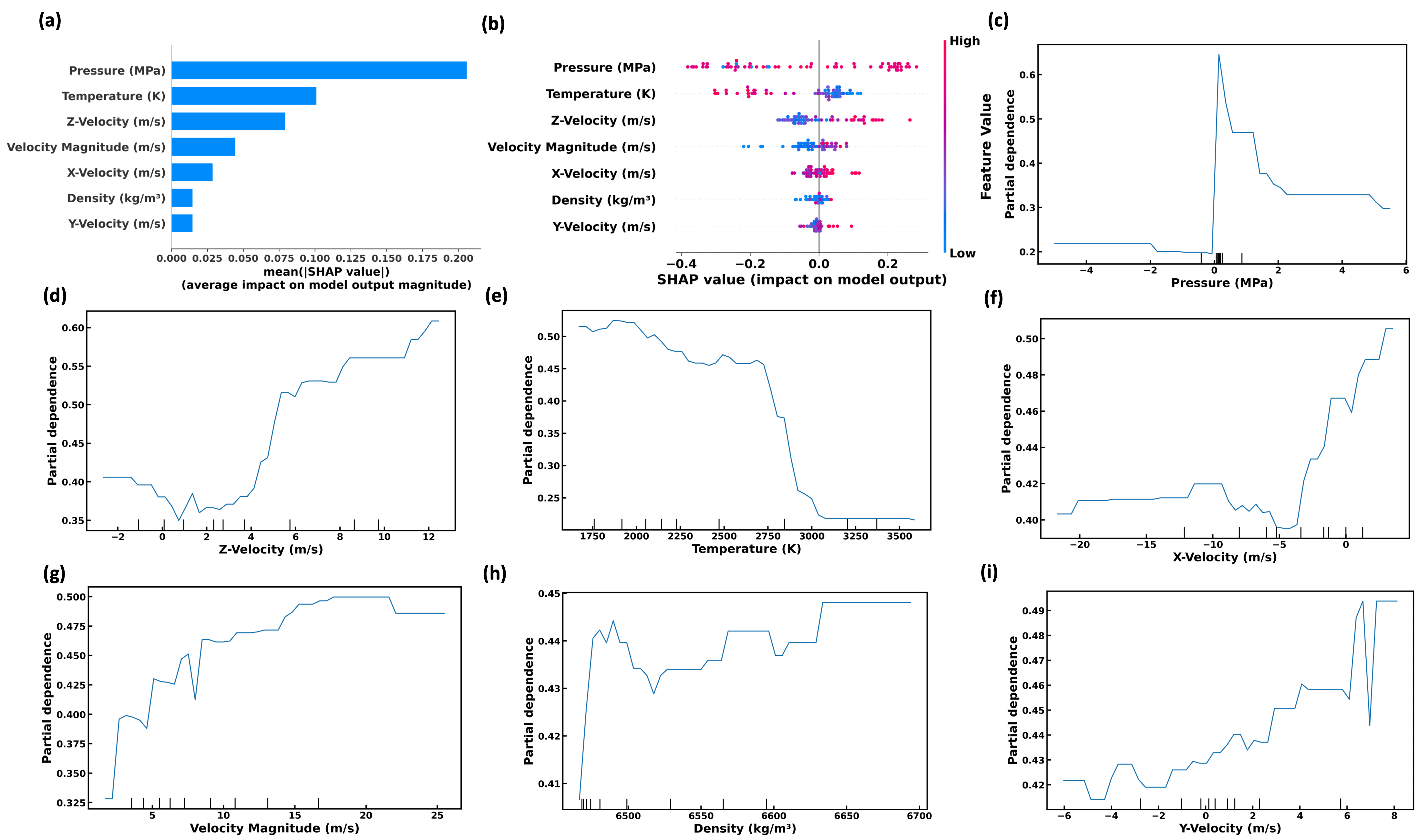}
\end{center}
    \caption{SHAP analysis on RF model in classifying spatter ejection and melt pool with the following features:  temperature, velocity, x-velocity, y-velocity, z-velocity, density and pressure (a) SHAP feature important (b) Balanced\_Accuracy score (c) z-axis PDP (d) pressure PDP (e) Partial dependence plot on y-axis PDP (f) z-velocity PDP (g) Velocity magnitude PDP (h) Temperature PDP (i) x-velocity PDP (j)  x-axis PDP (k) Density PDP(l)  y-velocity PDP }
\label{fig:Shap}
\end{figure*}

\subsection{ Spatter and Melt Pool Property Correlations}
The behavior of the melt pool depends on the process parameters, and while the ejected spatter originates from the melt pool, likely its properties are different. However, the exact nature and extent of these differences remain unknown as a result of the lack of equipment with the precision and sensitivity to detect such subtle variations. Collecting measurements during the vapor plume generation from the melt pool further complicates the data extraction because of the poor visibility and short timescale between changing melt pool dynamics and spatter ejections. Harnessing the strength of high-fidelity 3D simulation and ML models, the slight variation in the properties can appear in patterns across measurements and be detected; aiding in differentiating the measured properties of both the melt pool and the spatter at ejections. Statistical distributions were generated to demonstrate this effect, as shown in Figures \ref{fig:pairplot_coordinate} and \ref{fig:pairplot}. The diagonal of the pairplot displays a plotted histogram showing the distribution of the measured property. On the left-hand side of the diagonal are the kernel density estimate (KDE) plots showing the data density concentration, while on the right-hand side of the diagonal histogram are the scatter plots of the distribution. 
The differential in spatial location of the spatter and the melt pool depends on the ejection velocity and the simulation timestep at which it is captured. Compared to other spatial components (x and z axes),
a wider distribution was observed with respect to the y axis of the spatter than that of the melt pool in the histogram (yy plot), scatter plots (xy plot, yz plot) and KDE plots (xy plot, yz plot) as shown in Figure \ref{fig:pairplot_coordinate}. The spread of the spatter is slightly greater along the x-axis compared to the melt pool. Although the melt pool may exhibit a bow-like shape, leading to variations in height at different sampling locations, most spatter particles are still detected at higher altitudes, as shown in the scatter plots (xz plot and yz plot) and KDE plots (xz plot and yz plot) in Figure \ref{fig:pairplot_coordinate}. It is evident that the melt pool shows less variation along the y-axis (its width) compared to the x-axis (its length) and z-axis (its depth) during laser traversal across the powder bed.

 For the velocity distributions (x-velocity, y-velocity, z-velocity and velocity) shown in Figure \ref{fig:Openfoam}, the KDE spatter plots, the histogram, and the scatter plots of the spatter are wider than those of the melt pool, suggesting a wider range of velocities. The melt pool plots are narrower, indicating more uniform velocities. x-velocity distributions appear to be much closer between the spatter and the velocity, suggesting that the x-velocity might be a weak parameter in driving the spatter ejection. In addition, the density distributions between the spatter and the melt pool appear to be close to each other, which can be attributed to both the melt pool and the spatter liquid form. 

 In contrast, wider distribution spreads for the temperature and pressure of the melt pool were observed compared to those observed for the spatter, implying a broader temperature and pressure range of the melt pool. Spatter shows higher peaks in the histogram plots, indicating that the temperatures are more consistent and concentrated around a mean value. The scatter plots for the spatter display more defined groupings, while the melt pool's points are dispersed over a wider range, confirming the trends suggested by the histogram plots. The melt pool pressure distribution is wider also, as indicated by its histogram and scatter plot, whereas the spatter exhibits a sharper peak. This suggests a narrow range of pressure values could be required for spatter ejections, assuming that the melt pool's pressure is close to the spatter's pressure at the point of ejection via the evidence of the overlap of the pressure of both the melt pool and the spatter on the histogram plot. The scatter plots corresponding to the pressure of the melt pool against other parameters show a more dispersed pattern. The wider temperature and pressure distribution of the melt-pool data set is due to the high gradients of these properties along the melt-pool volume. Simultaneously, the narrower distribution of the spatter's temperature and pressure along with the reduced concentration at the upper bound compared to the melt pool could be due to its ejection occurring away from the laser impact on the material, resulting in cooling. The magnitude of cooling depends on the velocity of the ejection, especially when it travels away from the laser. In addition, the ejected spatter experiences pressure forces in all directions in the gas phase, causing an increase in the surface tension and the sphericity of their shape.   
 Overall, the spatter represents a more variable and less controlled process with a broader range of values across most of the parameters, whereas the melt pool indicates a more stable and consistent process where the parameters are more narrowly distributed around certain central values. Spatter ejection from the melt pool could be attributed to the locations in the melt pool that possess the observed magnitude or threshold for the combination of parameters. However, it is crucial to quantify these parameters based on their importance to spatter generation.  
\begin{figure*}[htbp!]
\begin{center}
\includegraphics[width=1\linewidth]{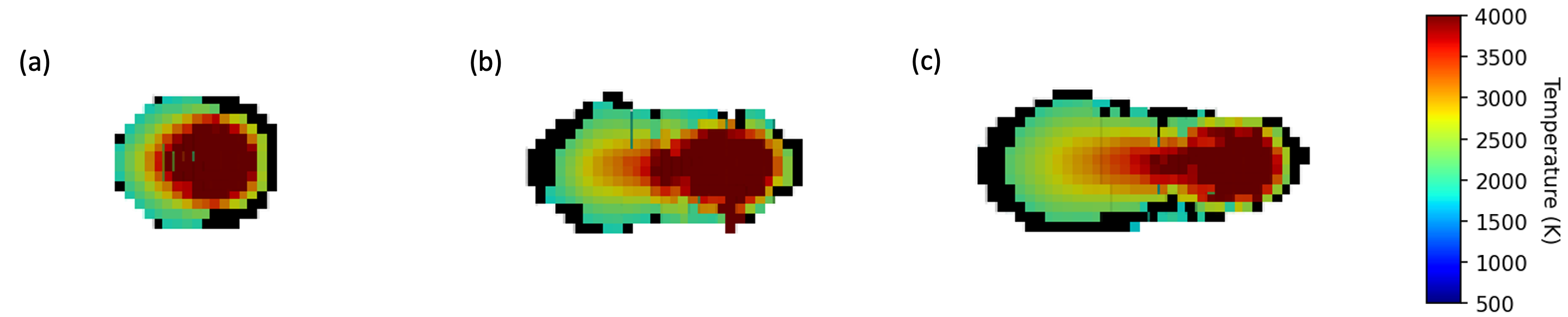}
\end{center}
    \caption{Spatter prediction on FLOW-3D simulation generated from testing FLOW-3D dataset at 400 W $2~\mathrm{ms}^{-1}$ on ML model that was trained using OpenFOAM dataset. Region masked with black are identified as location predicted as spatter (a) prediction at timestep 65 $\mu s$ (b)  prediction at timestep 165 $\mu s$  (c)  prediction at timestep 265 $\mu s$.}
\label{fig:FLOW3DPrediction}
\end{figure*}
\subsection{ Spatter and Melt Pool Classification via ML Model.}

  To investigate the performance of the OpenFOAM dataset on ML models, features including the x-axis, y-axis, z-axis, temperature, velocity, x-velocity, y-velocity, z-velocity, density, and pressure are used as the model input. Consistently high accuracy was observed across all ML models, with training accuracy ranging from 99\% to 100 \% and testing accuracy ranging from 94\% to 98\%. The high accuracy was maintained across all calculated score metrics (Accuracy, F1, Balanced Accuracy, ROC\_AUC) as shown in Figure \ref{fig:Corrdinate_Accuracy}(a-d) The lowest test accuracy of 94\% was observed with the KNN models. The closeness of the training and testing accuracy $< 0.1$)indicates that the ML models are neither under- or over-fitting.
The importance of the characteristics used in the ML test, as shown in \ref{fig:Corrdinate_Accuracy}e, followed the trend: z-axis $>$ pressure $>$ z-velocity $>$ y-axis $>$ velocity $>$ temperature $>$ x-velocity $>$ x-axis $<$ y-velocity $>$ density. The highest contribution of the z-axis to accuracy is attributed to the wide separation observed between the spatter and the melt pool when the z-axis is paired together with itself, or other parameters, as shown on the histogram, KDE, and scatter plots as shown in \ref{fig:pairplot_coordinate}.

The methodologies of our data extraction and combination of parameters for this classification task can serve as a benchmark in predicting spatter or developing reduced order model package for spatter generation either with desired simulation packages or collected experimental measurements. Use case of this technique is extended to simulation dataset generated with FLOW-3D. Modeling  LPBF in FLOW-3D generates a dataset that is well-suited for analyzing the melt pool rather than the spatter. Some parts of the melt pool dataset generated by FLOW-3D might have properties similar to those of spatter or meet the threshold required for spatter ejection. However, these locations remain part of the melt pool in the simulation because the physics module in FLOW-3D uses simplified assumptions for mass transfer between different phases, limiting its capability to accurately simulate spatter. Adapting the current ML model that was trained on OpenFOAM dataset on FLOW-3D dataset will necessitate removal of spatial components (x, y, z) from the features that will be used to train and test ML models. On training and testing OpenFOAM dataset without spatial components, high accuracies were still achieved with slight decrease by ~4\% across most of the model performances in comparison to when spatial components were present as shown Figure \ref{fig:Accuracy}. Significant decrease of 15 \% and 17 \% were observed with the SVC and KNN model respectively. The accuracy varies from 95 \% to 88 \%  for the RF and the bagging models respectively. High accuracy of the ML model with the features without spatial component qualify the RF model for predicting spatter on FLOW-3D package. To also ensure the models are not overfitting, both the training and testing score as shown in the Figure \ref{fig:Accuracy}. The largest difference between train and testing accuracy that was observed is  0.1 with respect to the bagging model, while the rest of the models are less than 0.05. These slight changes are small to consider the investigated model as not overfitting. The trend observed in the feature importance studies without spatial components, as shown in Figure \ref{fig:Accuracy}e follows the trend of pressure $>$ z-velocity $>$ temperature $>$ x-velocity $>$ velocity $>$ density $>$  y-velocity . Evidently, pressure is not only the most significant parameter driving the dynamics in the melt pool, but also indicate the possibility of spatter ejection at a magnitude that is confined within a narrow range as shown in \ref{fig:pairplot_coordinate}.  Strong importance observed with respect to z-velocity can be understood from the tendency of a body that possesses high z-velocity to travel upward which is associated with spatter, and we observed a narrow range for melt pool z-velocity, which is within the subset of the spatter's z-velocity. Also, observing temperature having less importance than the z-velocity despite its direct relationship with pressure according to Clausius–Clapeyron relation could be attributed to the saturation that is reached such that increment in the temperature will result in boiling of the liquid material which will limit the importance of temperature in promoting spatter ejection, rather causing more plume generation. In addition, temperature is the driving force that influences both recoil pressure and surface tension given that the material is fixed as it is in this study. Recoil pressure increases exponentially with temperature due to enhanced vaporization, resulting in stronger vapor thrust at high temperatures. In contrast, surface tension decreases linearly with increasing temperature, reflecting the weakening of cohesive forces at the molten metal surface. This inverse relationship suggests that at elevated temperatures, high recoil pressure promotes spatter formation, while low surface tension reduces the melt pool’s ability to maintain cohesive boundaries, leading to spatter ejection. For a  material with a  high surface tension coefficient, the resulting Marangoni flow will be stronger, potentially stabilizing the melt pool through efficient heat redistribution making  the flow become turbulent. In addition to the recoil pressure and surface tension, marangoni convection, buoyancy, surface tension, and gravity are forces that also define the local and global force balance within the melt pool \cite{alphonso2023possibility}. When this balance is disrupted due to excessive recoil, sharp Marangoni gradients, or thermal instabilities, the result is a highly dynamic melt pool with irregular spatter ejection, and increased risk of defects such as porosity or surface roughness \cite{tian2020numerical}.  High contribution of z-velocity to the spatter ejection could also be understood from the parabolic shape of the melt pool which is formed as the laser traverse in x-axis direction. Hot fluid will build up in the direction of the laser, which can break up to create spatter due to its elongation as shown in Figure 8a, via shear thinning and later break-up as shown in Figure 8b.  Additionally, spatter breakage can further intensify the backward dispersion of material and energy that are in form of a wave which can travel through the high gradient region of the melt pool, creating z-velocity component that could enhance ejection. The backward dispersion of material and energy in the melt pool can be attributed to the combination of other factors such as pressure, temperature gradient, and the vibration in the melt pool which can also influence other properties like viscosity and surface tension to cause Marangoni effect and promote melt pool flow, elongation and ultimately fluid ejections. To gain further insight to the accuracy of the ML model predictions, we constructed a precision recall table as shown on Figure \ref{fig:Accuracy}f and \ref{fig:Corrdinate_Accuracy}f for predictions with spatial components included in the features and with no spatial components respectively. We observed our model having the capability in predicting most of the points with the melt pool and the spatter. The wrong prediction observations for the model with the presence of spatial component features  is associated with regions that were melt pool but were falsely predicted as spatter. While the wrong prediction observations for the model with the absent of spatial component features  is associated with regions that were spatter but were falsely predicted as melt pool. 
\subsection{ Explainable AI via SHAP Analysis.}
The \texttt{shap.TreeExplainer} was used to calculate the SHAP values for the test dataset, focusing on the contribution of each feature to the prediction of the positive class (spatter). Comparison of SHAP importance and ML importance ensures consistency in feature relevance across different methods. A summary bar plot was generated to visualize the mean absolute SHAP values, which indicate the average magnitude of feature impacts on the model's output. The SHAP summary bar plot offers a high-level overview of the importance of the features. Figure \ref{fig:Shap_coordinate}a and Figure \ref{fig:Shap}a present the feature important from the SHAP analysis for the experiment with and without spatial components respectively, in the datasets. The x-axis represents the mean absolute SHAP values for each feature, indicating the average magnitude of a feature's impact on the model's output. Features are ranked in descending order, making it easy to identify the most influential features driving the predictions. The SHAP feature importance rankings are consistent with the feature importance rankings derived directly from the ML model in Figure \ref{fig:Corrdinate_Accuracy}e and Figure \ref{fig:Accuracy}e indicating consistency between the model's internal behavior and the SHAP explanation.  

The SHAP summary plot with a color bar adds depth by visualizing the distribution of SHAP values for individual samples. Each point corresponds to a sample, with its position along the x-axis representing its SHAP value and its color encoding the feature value (e.g., high or low). This plot reveals both the importance and directionality of features, indicating patterns and interactions in the data. The farther the SHAP values deviate from zero, the stronger the influence of the feature on the model output, which was observed to follow the trends observed in \ref{fig:Corrdinate_Accuracy}b and Figure \ref{fig:Accuracy}b. Furthermore, high values (red) are predominantly associated with positive SHAP values in high-influencing features such as z-axis, pressure, z-velocity, velocity magnitude, indicating their ability to increase the likelihood of spatter predictions. In reverse, a low value (red) of a variable such as temperature tends to positively contribute to spatter generation.

To better understand the effect of the variation in the value of a particular feature on the prediction of the model, PDP plots were generated for each feature as shown on the Figure \ref{fig:Shap_coordinate}(c-l) and Figure \ref{fig:Shap}(c-i). The y-axis of the PDP shows the average predicted probability of the positive class (spatter). In Figure \ref{fig:Shap_coordinate}c, we observed that the z-axis value has a significant threshold effect from  -10 $\mu m$ to 40 $\mu m$ suggesting this range is critical for the model's predictions, while beyond ~100 $\mu m$, the impact saturates. It is likely that the timestep at which OpenFOAM simulation was conducted will affect the observed range in Z-Axis. The extreme negative and positive values of the Y-axis strongly influence the model output, while values from  -70 $\mu m$ to 50 $\mu m$ minimal effect as it appear to be steady. to influence the predictions of the model with respect to the pressure (Figure \ref{fig:Shap_coordinate}d) and  (Figure \ref{fig:Shap}c), below ~-0.5 MPa and above ~-2 MPa , the partial dependence remains flat, indicating negligible impact on predictions. A sharp increase in partial dependence is observed as the pressure increases from approximately 0 to 0.2 MPa, where a significant positive impact is observed on the model predictions. The partial dependence is nearly flat and stable around 0.35, indicating that pressure in this range has minimal influence on the model's predictions. there was a decrease from 1.6 MPa to 2 MPa, which slightly favors spatter ejection. The curve plateaus around 2 MPa to 5.8 MPa, suggesting that increasing pressure beyond this point does not have a substantial additional impact on the predictions. A significant increase in z-velocity was observed from $2~\mathrm{ms}^{-1}$ to $5~\mathrm{ms}^{-1}$ (Figure \ref{fig:Shap_coordinate}f), while the velocity magnitude initially increased at $2.5~\mathrm{ms}^{-1}$, then almost stabilized to $10~\mathrm{ms}^{-1}$ where it continued to increase to $17~\mathrm{ms}^{-1}$ (Figure \ref{fig:Shap_coordinate} g). Close pattern was observed in the PDP plot of features with the ML experiment, where there are no spatial components in the input features in Figure \ref{fig:Shap}.
Steady reduction in temperature was observed to favor spatter generation (Figure \ref{fig:Shap_coordinate}h and Figure \ref{fig:Shap}e) which could be attributed to spatter generation occurring mostly away from the location of the laser, as the fluid swirl backward due to the Marangoni recirculation effect. Along the x-axis (the laser traveling direction) in Figure \ref{fig:Shap_coordinate}j, SHAP value decline at the initial stage then remain almost steady at the higher value X-Axis. Although the SHAP values of features like Y-Velocity, density, and X-Axis are small, it is likely that they promote the accuracy predictions in our studies due to their tendency to interact with features with high SHAP values. Khairallah et al. \cite{khairallah2016laser} demonstrated the critical role of recoil pressure and Marangoni convection in shaping melt pool flow and influencing the formation of defects such as spattering, denudation, and pores. They found that  as temperature rises, the recoil pressure increases, overcoming surface tension and generating depressions and spatter. Expanding from the findings from Khairallah et al. \cite{khairallah2016laser} on the investigation of spatter, by coupling multi-physics and ML techniques, ranges at which different parameter leads to spatter formation is unveiled.

\subsection{ Spatter Prediction on a Commercial Package "FLOW-3D".}
Spatter prediction on FLOW-3D is promoted by two factors: (1) the close alignment between the characteristics of the melting pool dimensions and the temperature distribution of OpenFOAM and FLOW-3D and (2) the high precision of the ML model in classifying the spatter from the melting pool. 
Upon testing our model on FLOW-3D, voxels that match the spatter property were identified and masked with black points as shown in Figure \ref{fig:FLOW3DPrediction}. At the early timesteps of the FLOW-3D simulation, spatter predictions were predominantly observed at the front of the melt pool (Figure \ref{fig:FLOW3DPrediction}a), then at the last time steps (Figure \ref{fig:FLOW3DPrediction}b and Figure \ref{fig:FLOW3DPrediction}c), the spatter tends to occupy the backward side of the melt pool,  this is similar to the observed OpenFOAM simulation for spatter generation as shown in Figure \ref{fig:Openfoam} and the experimental result reported by Yin et al. \cite{yin2020correlation}. Note that conditions for spatter generation were not met at the location of the hotspot of the melt pool because of the high temperature that could be causing fluid evaporation. Consequently, the high accuracy outcome of the Random Forest (RF) model underscores the efficacy of our developed methodology for the reduced order task. This methodology can enhance the robustness of commercial packages by synergizing them with machine learning (ML) models to account for complex and computationally expensive phenomena, such as spatter modeling. An alternative application of the developed methodology is testing datasets of different process parameters generated from commercial software, such as FLOW-3D, towards the development of a process map.

\subsection{Spatter Process Map}
\begin{figure*}[htbp!]
\begin{center}
\includegraphics[width=0.8\linewidth]{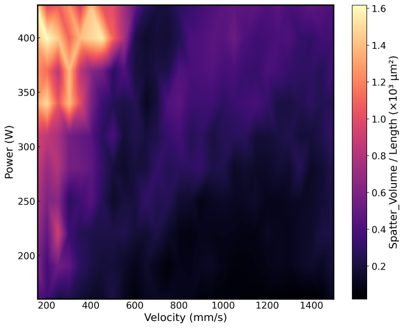}
\end{center}
    \caption{Process map generated via hybrid of OpenFOAM trained ML model on FLOW-3D dataset to understand relationship between spatter formation and process conditions. The process map was created using 281 FLOW-3D simulations with different combinations of power and velocity.}
\label{fig:ProcessMAP}
\end{figure*}

Display of spatter quantification with varying power/velocity combinations will provide access to a suitable operating window with minimal defects for part production. Running OpenFOAM for additive manufacturing is computationally expensive due to a lack of simplifying assumptions in the physics it is configured to solve as compared to FLOW-3D, that has a devoted module for AM. On the same machine, both FLOW-3D and OpenFOAM require run times of 4 and 72 hours respectively to generate LPBF solution on 5 $\mu m$ mesh size using Intel(R)Xeon(R) Gold 6230R CPU @ 2.10GHz, 2095Mhz, 26 cores. Building a hybrid of OpenFOAM and FLOW-3D via the high accuracy ML model as a reduced order model is attractive for large dataset generation for the development of spatter process map. Figure \ref{fig:ProcessMAP} is generated by testing 281 dataset consisting varying power/velocity combinations of FLOW-3D experiments on OpenFOAM trained ML model. 
As shown in Figure \ref{fig:ProcessMAP}, the spatter quantification follows a predictable trend within processing parameter space for LPBF, except in a region between 400 - 800 $mm s^{-1}$ where spatter count is consistently low across the power range investigated. This is also clearly demonstrated in Figures A.11 and A.12 of the Appendix. It was observed that higher spatter was generated at high power and low velocity, which is the region associated with keyhole formation. Conversely, at low power and high speed, less spatter was observed, corresponding to the region characterized by lack of fusion \cite{gordon2020defect}. Based on the process conditions, we observed the trend followed by the volume of spatter generation as  high power/low speed $>$ high power/high speed $>$ low power/low speed $>$ low power/high speed. The observed trend can be attributed to the intensity of the interaction between the material and the laser, leading to variations in melt pool size, and the magnitude of mass and energy transport. Consequently, high power and low velocity produce the largest melt pool with the most intense transport phenomena and kinetic effects, resulting in the highest spatter ejection. In contrast, low power and high velocity create the smallest melt pool with the least intense effects, resulting in minimal spatter ejection. Certain locations within the safe operating window for porosity (situated between keyhole formation and lack of fusion) \cite{gordon2020defect} intersect with areas of low spatter count, as shown in \ref{fig:ProcessMAP}. These intersections make those regions promising for part printing. 
The reason for the observed low spatter volume at a velocity between 400 - 800 $mm/s$ throughout the investigated power range is probably the transition between a kinetic regime and a diffusion-limited regime resulting from a change in the morphology of the melt pool.  This transition creates a balanced environment in which the dynamics of the melt pool is stabilized, leading to reduced ejection of spatters due to moderate vaporization, stable formation of the melt pool, and efficient heat distribution.
The analysis of the study by Zheng et al.~\cite{zheng2018effects}
is consistent with the low-spatter band observed on our process map
between 400 and 800~mm/s. They examined the formation and evolution of
the vapor plume and its effect on spatter generation in stainless steel
at scan speeds of 200, 400, 500, 1000, and 2000~mm/s, and reported that
spatter generation depends strongly on plume stability. At the lowest
speed (200~mm/s), the high energy density produced a highly unstable
plume, with recoil pressure strong enough to eject droplets from the
melt pool. As the speed increased to 400--500~mm/s, spattering and plume
formation barely occurred, in agreement with the reduced spatter counts
in our band. Increasing the speed further to 1000 and 2000~mm/s
destabilized the melt track, producing periodic swelling connected by
narrow bottlenecks of ripples along the track, a result of the
Plateau--Rayleigh capillary instability~\cite{khairallah2014mesoscopic}.
Within the upper portion of this speed range, the vapor plume tilts
backward relative to the scan direction, which is attributed to a shift
in the laser--material interaction interface from the bottom to the
front wall of the depression. The resulting backward vapor flux is
thought to influence the melt-pool fluid dynamics and increase the
susceptibility of the melt-track end to open-pore formation.

\section{Conclusion}
In this work, we developed a high-fidelity multiphysics simulation framework in OpenFOAM to model single-track selective laser melting and generate detailed 3D datasets capturing melt pool dynamics and hot spatter formation. A custom spatter-tracking algorithm was implemented to extract physically meaningful features for machine learning analysis. Different machine learning classification models, including XGBoost, LightGBM, Random Forest, Bagging, SVC, and KNN, were trained on this dataset. Among them, Random Forest achieved the highest accuracy of 94\% on the test dataset. The influence of the input features to the model was evaluated using feature importance and SHAP analysis, which revealed their contributions to spatter formation.
The trained machine learning models were further applied to FLOW-3D simulation outputs to identify spatter-prone regions on a voxel-by-voxel basis, demonstrating the feasibility of transferring OpenFOAM-trained classifiers to a computationally efficient simulation environment. Additionally, a process map was constructed to link laser power and scan speed to the predicted hot-spatter quantity based on the 281-case FLOW-3D dataset.

The present framework should be interpreted as a material-specific and hot-spatter-focused methodology. It does not constitute a complete predictive model for all spatter mechanisms in LPBF. In particular, powder entrainment, cold spatter, particle agglomeration, and material-to-material variations in surface tension, recoil pressure, viscosity, absorptivity, and Marangoni behavior remain outside the scope of the current work. Future studies will extend this framework using DEM--CFD coupling, additional material-specific descriptors, and multi-class machine learning models to distinguish hot, cold, powder-driven, and agglomerated spatter events. 

\section*{Declarations}
\subsection*{Funding}

This research was sponsored by the Army Research Laboratory and conducted under Cooperative Agreement Number W911NF-20-2-0175. The views and conclusions presented in this document are those of the authors and do not necessarily represent the official policies or positions of the Army Research Laboratory or the U.S. Government. The U.S. Government is authorized to reproduce and distribute reprints for governmental purposes, despite any copyright notice herein.

\subsection*{Competing Interests}

The authors have no competing interests to declare that are relevant to the content of this article.

\bibliography{reference}

@article{wirth2022implementation,
  title={Implementation of the Marangoni effect in an open-source software environment and the influence of surface tension modeling in the mushy region in laser powder bed fusion (LPBF)},
  author={Wirth, Florian and Tonn, Teresa and Sch{\"o}berl, Markus and Hermann, Stefan and Birkhofer, Hannes and Ploshikhin, Vasily},
  journal={Modelling and Simulation in Materials Science and Engineering},
  volume={30},
  number={3},
  pages={034001},
  year={2022},
  publisher={IOP Publishing}
}

@article{galler1964improved,
  title={An improved equivalence algorithm},
  author={Galler, Bernard A and Fisher, Michael J},
  journal={Communications of the ACM},
  volume={7},
  number={5},
  pages={301--303},
  year={1964},
  publisher={ACM New York, NY, USA}
}

@article{scikit-image,
 title = {scikit-image: image processing in {P}ython},
 author = {van der Walt, {S}t\'efan and {S}ch\"onberger, {J}ohannes {L}. and
           {Nunez-Iglesias}, {J}uan and {B}oulogne, {F}ran\c{c}ois and {W}arner,
           {J}oshua {D}. and {Y}ager, {N}eil and {G}ouillart, {E}mmanuelle and
           {Y}u, {T}ony and the scikit-image contributors},
 year = {2014},
 month = {6},
 volume = {2},
 pages = {e453},
 journal = {PeerJ},
 issn = {2167-8359},
 url = {https://doi.org/10.7717/peerj.453},
 doi = {10.7717/peerj.453}
}

@article{fiorio1996two,
  title={Two linear time union-find strategies for image processing},
  author={Fiorio, Christophe and Gustedt, Jens},
  journal={Theoretical Computer Science},
  volume={154},
  number={2},
  pages={165--181},
  year={1996},
  publisher={Elsevier}
}

@article{ogoke2023convolutional,
  title={Convolutional neural networks for melt depth prediction and visualization in laser powder bed fusion},
  author={Ogoke, Francis and Lee, William and Kao, Ning-Yu and Myers, Alexander and Beuth, Jack and Malen, Jonathan and Barati Farimani, Amir},
  journal={The International Journal of Advanced Manufacturing Technology},
  volume={129},
  number={7},
  pages={3047--3062},
  year={2023},
  publisher={Springer}
}

@article{myers2023high,
  title={High-resolution melt pool thermal imaging for metals additive manufacturing using the two-color method with a color camera},
  author={Myers, Alexander J and Quirarte, Guadalupe and Ogoke, Francis and Lane, Brandon M and Uddin, Syed Zia and Farimani, Amir Barati and Beuth, Jack L and Malen, Jonathan A},
  journal={Additive Manufacturing},
  volume={73},
  pages={103663},
  year={2023},
  publisher={Elsevier}
}

@article{hemmasian2023surrogate,
  title={Surrogate modeling of melt pool temperature field using deep learning},
  author={Hemmasian, AmirPouya and Ogoke, Francis and Akbari, Parand and Malen, Jonathan and Beuth, Jack and Farimani, Amir Barati},
  journal={Additive Manufacturing Letters},
  volume={5},
  pages={100123},
  year={2023},
  publisher={Elsevier}
}

@article{repossini2017use,
  title={On the use of spatter signature for in-situ monitoring of Laser Powder Bed Fusion},
  author={Repossini, Giulia and Laguzza, Vittorio and Grasso, Marco and Colosimo, Bianca Maria},
  journal={Additive Manufacturing},
  volume={16},
  pages={35--48},
  year={2017},
  publisher={Elsevier}
}

@article{cunningham2017synchrotron,
  title={Synchrotron-based X-ray microtomography characterization of the effect of processing variables on porosity formation in laser power-bed additive manufacturing of Ti-6Al-4V},
  author={Cunningham, Ross and Narra, Sneha P and Montgomery, Colt and Beuth, Jack and Rollett, AD},
  journal={Jom},
  volume={69},
  pages={479--484},
  year={2017},
  publisher={Springer}
}

@article{young2020types,
  title={Types of spatter and their features and formation mechanisms in laser powder bed fusion additive manufacturing process},
  author={Young, Zachary A and Guo, Qilin and Parab, Niranjan D and Zhao, Cang and Qu, Minglei and Escano, Luis I and Fezzaa, Kamel and Everhart, Wes and Sun, Tao and Chen, Lianyi},
  journal={Additive Manufacturing},
  volume={36},
  pages={101438},
  year={2020},
  publisher={Elsevier}
}

@article{anwar2018study,
  title={Study of the spatter distribution on the powder bed during selective laser melting},
  author={Anwar, Ahmad Bin and Pham, Quang-Cuong},
  journal={Additive Manufacturing},
  volume={22},
  pages={86--97},
  year={2018},
  publisher={Elsevier}
}

@article{wang2020investigation,
  title={Investigation into Spatter Particles and Their Effect on the Formation Quality During Selective Laser Melting Processes.},
  author={Wang, Zhiqiang and Wang, Xuede and Zhou, Xin and Ye, Guangzhao and Cheng, Xing and Zhang, Peiyu},
  journal={CMES-Computer Modeling in Engineering \& Sciences},
  volume={124},
  number={1},
  year={2020}
}

@article{hojjatzadeh2020direct,
  title={Direct observation of pore formation mechanisms during LPBF additive manufacturing process and high energy density laser welding},
  author={Hojjatzadeh, S Mohammad H and Parab, Niranjan D and Guo, Qilin and Qu, Minglei and Xiong, Lianghua and Zhao, Cang and Escano, Luis I and Fezzaa, Kamel and Everhart, Wes and Sun, Tao and others},
  journal={International Journal of Machine Tools and Manufacture},
  volume={153},
  pages={103555},
  year={2020},
  publisher={Elsevier}
}

@article{martin2019dynamics,
  title={Dynamics of pore formation during laser powder bed fusion additive manufacturing},
  author={Martin, Aiden A and Calta, Nicholas P and Khairallah, Saad A and Wang, Jenny and Depond, Phillip J and Fong, Anthony Y and Thampy, Vivek and Guss, Gabe M and Kiss, Andrew M and Stone, Kevin H and others},
  journal={Nature communications},
  volume={10},
  number={1},
  pages={1987},
  year={2019},
  publisher={Nature Publishing Group UK London}
}

@article{mukherjee2018mitigation,
  title={Mitigation of lack of fusion defects in powder bed fusion additive manufacturing},
  author={Mukherjee, T and DebRoy, Tarasankar},
  journal={Journal of Manufacturing Processes},
  volume={36},
  pages={442--449},
  year={2018},
  publisher={Elsevier}
}

@article{promoppatum2022quantification,
  title={Quantification and prediction of lack-of-fusion porosity in the high porosity regime during laser powder bed fusion of Ti-6Al-4V},
  author={Promoppatum, Patcharapit and Srinivasan, Raghavan and Quek, Siu Sin and Msolli, Sabeur and Shukla, Shashwat and Johan, Nur Syafiqah and van der Veen, Sjoerd and Jhon, Mark Hyunpong},
  journal={Journal of Materials Processing Technology},
  volume={300},
  pages={117426},
  year={2022},
  publisher={Elsevier}
}

@article{yin2020correlation,
  title={Correlation between forming quality and spatter dynamics in laser powder bed fusion},
  author={Yin, Jie and Wang, Dengzhi and Yang, Liangliang and Wei, Huiliang and Dong, Peng and Ke, Linda and Wang, Guoqing and Zhu, Haihong and Zeng, Xiaoyan},
  journal={Additive Manufacturing},
  volume={31},
  pages={100958},
  year={2020},
  publisher={Elsevier}
}

@inproceedings{barrett2018low,
  title={Low cost, high speed stereovision for spatter tracking in laser powder bed fusion},
  author={Barrett, Christopher and Carradero, Carolyn and Harris, Evan and McKnight, Jeremy and Walker, Jason and MacDonald, Eric and Conner, Brett},
  booktitle={2018 International Solid Freeform Fabrication Symposium},
  year={2018},
  organization={University of Texas at Austin}
}

@article{barrett2019statistical,
  title={Statistical analysis of spatter velocity with high-speed stereovision in laser powder bed fusion},
  author={Barrett, Christopher and Carradero, Carolyn and Harris, Evan and Rogers, Kirk and MacDonald, Eric and Conner, Brett},
  journal={Progress in Additive Manufacturing},
  volume={4},
  number={4},
  pages={423--430},
  year={2019},
  publisher={Springer}
}

@article{deng2019investigation,
  title={Investigation of spatter occurrence in remote laser spiral welding of zinc-coated steels},
  author={Deng, Shengjie and Wang, Hui-Ping and Lu, Fenggui and Solomon, Joshua and Carlson, Blair E},
  journal={International Journal of Heat and Mass Transfer},
  volume={140},
  pages={269--280},
  year={2019},
  publisher={Elsevier}
}

@article{yang2019predictive,
  title={Predictive large-eddy-simulation wall modeling via physics-informed neural networks},
  author={Yang, XIA and Zafar, Suhaib and Wang, J-X and Xiao, Heng},
  journal={Physical Review Fluids},
  volume={4},
  number={3},
  pages={034602},
  year={2019},
  publisher={APS}
}

@article{du2021physics,
  title={Physics-informed machine learning and mechanistic modeling of additive manufacturing to reduce defects},
  author={Du, Yang and Mukherjee, Tuhin and DebRoy, Tarasankar},
  journal={Applied Materials Today},
  volume={24},
  pages={101123},
  year={2021},
  publisher={Elsevier}
}

@article{wang2020machine,
  title={Machine learning in additive manufacturing: State-of-the-art and perspectives},
  author={Wang, Chengcheng and Tan, XP and Tor, Shu Beng and Lim, CS},
  journal={Additive Manufacturing},
  volume={36},
  pages={101538},
  year={2020},
  publisher={Elsevier}
}

@article{stathatos2019real,
  title={Real-time simulation for long paths in laser-based additive manufacturing: a machine learning approach},
  author={Stathatos, Emmanuel and Vosniakos, George-Christopher},
  journal={The International Journal of Advanced Manufacturing Technology},
  volume={104},
  pages={1967--1984},
  year={2019},
  publisher={Springer}
}

@article{akbari2022meltpoolnet,
  title={MeltpoolNet: Melt pool characteristic prediction in Metal Additive Manufacturing using machine learning},
  author={Akbari, Parand and Ogoke, Francis and Kao, Ning-Yu and Meidani, Kazem and Yeh, Chun-Yu and Lee, William and Farimani, Amir Barati},
  journal={Additive Manufacturing},
  volume={55},
  pages={102817},
  year={2022},
  publisher={Elsevier}
}

@article{jadhav2023stressd,
  title={StressD: 2D Stress estimation using denoising diffusion model},
  author={Jadhav, Yayati and Berthel, Joseph and Hu, Chunshan and Panat, Rahul and Beuth, Jack and Farimani, Amir Barati},
  journal={Computer Methods in Applied Mechanics and Engineering},
  volume={416},
  pages={116343},
  year={2023},
  publisher={Elsevier}
}

@article{ren2023machine,
  title={Machine learning--aided real-time detection of keyhole pore generation in laser powder bed fusion},
  author={Ren, Zhongshu and Gao, Lin and Clark, Samuel J and Fezzaa, Kamel and Shevchenko, Pavel and Choi, Ann and Everhart, Wes and Rollett, Anthony D and Chen, Lianyi and Sun, Tao},
  journal={Science},
  volume={379},
  number={6627},
  pages={89--94},
  year={2023},
  publisher={American Association for the Advancement of Science}
}

@article{ly2017metal,
  title={Metal vapor micro-jet controls material redistribution in laser powder bed fusion additive manufacturing},
  author={Ly, Sonny and Rubenchik, Alexander M and Khairallah, Saad A and Guss, Gabe and Matthews, Manyalibo J},
  journal={Scientific reports},
  volume={7},
  number={1},
  pages={4085},
  year={2017},
  publisher={Nature Publishing Group UK London}
}

@article{gordon2020defect,
  title={Defect structure process maps for laser powder bed fusion additive manufacturing},
  author={Gordon, Jerard V and Narra, Sneha P and Cunningham, Ross W and Liu, He and Chen, Hangman and Suter, Robert M and Beuth, Jack L and Rollett, Anthony D},
  journal={Additive Manufacturing},
  volume={36},
  pages={101552},
  year={2020},
  publisher={Elsevier}
}

@article{RF,
title  = {Random Forests.},
author = {Breiman, L.},
journal  = {Machine Learning},
year  = {2001},
volume  = {45},
issue  = {},
pages  = {5–32},
doi  = {10.1023/A:1010933404324},
}

@article{Ridge,
author = { Arthur E.   Hoerl  and  Robert W.   Kennard },
title = {Ridge Regression: Biased Estimation for Nonorthogonal Problems},
journal = {Technometrics},
volume = {12},
number = {1},
pages = {55-67},
year  = {1970},
publisher = {Taylor & Francis},
doi = {10.1080/00401706.1970.10488634},
}

@article{geurts2006extremely,
  title={Extremely randomized trees},
  author={Geurts, Pierre and Ernst, Damien and Wehenkel, Louis},
  journal={Machine learning},
  volume={63},
  number={1},
  pages={3--42},
  year={2006},
  publisher={Springer}
}

@inproceedings{chen2016xgboost,
  title={Xgboost: A scalable tree boosting system},
  author={Chen, Tianqi and Guestrin, Carlos},
  booktitle={Proceedings of the 22nd acm sigkdd international conference on knowledge discovery and data mining},
  pages={785--794},
  year={2016}
}

@inproceedings{ke2017lightgbm,
  title={Lightgbm: A highly efficient gradient boosting decision tree},
  author={Ke, Guolin and Meng, Qi and Finley, Thomas and Wang, Taifeng and Chen, Wei and Ma, Weidong and Ye, Qiwei and Liu, Tie-Yan},
  booktitle={Advances in neural information processing systems},
  pages={3146--3154},
  year={2017}
}

@article{breiman2001random,
  title={Random forests},
  author={Breiman, Leo},
  journal={Machine learning},
  volume={45},
  number={1},
  pages={5--32},
  year={2001},
  publisher={Springer}

}

@article{breiman1996bagging,
  title={Bagging predictors},
  author={Breiman, Leo},
  journal={Machine learning},
  volume={24},
  number={2},
  pages={123--140},
  year={1996},
  publisher={Springer}
}

@article{cortes1995support,
  title={Support-vector networks},
  author={Cortes, Corinna and Vapnik, Vladimir},
  journal={Machine learning},
  volume={20},
  number={3},
  pages={273--297},
  year={1995},
  publisher={Springer}
}

@Manual{FLOW-3D,
author = {Flow Science, Inc.},

title = {FLOW-3D, Version~12.0},

year = {2019},

address = {Santa Fe, NM},

url = {https://www.flow3d.com/}

}

@article{ogoke2023inexpensive,
  title={Inexpensive High Fidelity Melt Pool Models in Additive Manufacturing Using Generative Deep Diffusion},
  author={Ogoke, Francis and Liu, Quanliang and Ajenifujah, Olabode and Myers, Alexander and Quirarte, Guadalupe and Beuth, Jack and Malen, Jonathan and Farimani, Amir Barati},
  journal={arXiv preprint arXiv:2311.16168},
  year={2023}
}

@article{sheikh2024exploring,
  title={Exploring chemistry and additive manufacturing design spaces: a perspective on computationally-guided design of printable alloys},
  author={Sheikh, Sofia and Vela, Brent and Attari, Vahid and Huang, Xueqin and Morcos, Peter and Hanagan, James and Acemi, Cafer and Karaman, Ibrahim and Elwany, Alaa and Arroyave{\'{}}, Raymundo},
  journal={Materials Research Letters},
  volume={12},
  number={4},
  pages={235--263},
  year={2024},
  publisher={Taylor \& Francis}
}

@article{yao2017hybrid,
  title={A hybrid machine learning approach for additive manufacturing design feature recommendation},
  author={Yao, Xiling and Moon, Seung Ki and Bi, Guijun},
  journal={Rapid Prototyping Journal},
  volume={23},
  number={6},
  pages={983--997},
  year={2017},
  publisher={Emerald Publishing Limited}
}

@article{bendsoe2005topology,
  title={Topology optimization-broadening the areas of application},
  author={Bendsoe, Martin P and Lund, Erik and Olhoff, Niels and Sigmund, Ole},
  journal={Control and Cybernetics},
  volume={34},
  number={1},
  pages={7--35},
  year={2005},
  publisher={Polska Akademia Nauk. Instytut Bada{\'n} Systemowych PAN}
}

@article{panwisawas2020metal,
  title={Metal 3D printing as a disruptive technology for superalloys},
  author={Panwisawas, Chinnapat and Tang, Yuanbo T and Reed, Roger C},
  journal={Nature Communications},
  volume={11},
  number={1},
  pages={2327},
  year={2020},
  publisher={Nature Publishing Group UK London}
}

@article{li2022review,
  title={A review of spatter in laser powder bed fusion additive manufacturing: In situ detection, generation, effects, and countermeasures},
  author={Li, Zheng and Li, Hao and Yin, Jie and Li, Yan and Nie, Zhenguo and Li, Xiangyou and You, Deyong and Guan, Kai and Duan, Wei and Cao, Longchao and others},
  journal={Micromachines},
  volume={13},
  number={8},
  pages={1366},
  year={2022},
  publisher={MDPI}
}

@misc{msmsac4a26bib17,
  author={OpenCFD Ltd (ESI Group)},
  title={OpenFOAM documentation},
  note={(online) https://www.openfoam.com/documentation/overview)}
}

@misc{msmsac4a26bib8,
  author={OpenCFD Ltd (ESI Group)},
  title={OpenFOAM news},
  note={(online) https://www.openfoam.com/news/main-news/openfoam-v20-12)}
}

@article{msmsac4a26bib15,
  author={J U Brackbill and D B Kothe and C Zemach},
  title={A continuum method for modeling surface tension},
  journal={J. Comput. Phys.},
  volume={100},
  year={1992},
  pages={335--354},
  url={https://doi.org/10.1016/0021-9991(92)90240-y},
  DOI={10.1016/0021-9991(92)90240-y}
}

@inproceedings{wirth2021influence,
  title={Influence of the inert gas flow on the laser powder bed fusion (LPBF) process},
  author={Wirth, Florian and Frauchiger, Alex and Gutknecht, Kai and Cloots, Michael},
  booktitle={Industrializing Additive Manufacturing: Proceedings of AMPA2020},
  pages={192--204},
  year={2021},
  organization={Springer}
}

@misc{lundkvist2019cfd,
  title={CFD Simulation of Fluid Flow During Laser Metal Wire Deposition using OpenFOAM: 3D printing},
  author={Lundkvist, Jennifer},
  year={2019}
}

@article{bahreini2016development,
  title={Development of a phase change model for volume-of-fluid method in OpenFOAM},
  author={Bahreini, Mohammad and Ramiar, Abbas and Ranjbar, Ali Akbar},
  journal={Journal of Heat and Mass Transfer Research},
  volume={3},
  number={2},
  pages={131--143},
  year={2016},
  publisher={Semnan University Press}
}

@article{lee1980pressure,
  title={A pressure iteration scheme for two-phase flow modeling},
  author={Lee, Wen Ho},
  journal={Multiphase transport fundamentals, reactor safety, applications},
  volume={1},
  pages={407--431},
  year={1980},
  publisher={Washington, DC}
}

@inproceedings{scuro2022progress,
  title={Progress in Coupling Computational Thermodynamics and Computational Fluid Dynamics to Support Molten Salt Reactor Applications},
  author={Scuro, NL and Poschmann, M and Bene{\v{s}}, O and Piro, MHA},
  booktitle={Conference Paper: G4SR4},
  pages={1--15},
  year={2022}
}

@article{o2024computational,
  title={Computational analysis and experiments of spatter transport in a laser powder bed fusion machine},
  author={O’Brien, Nicholas and Uddin, Syed Zia and Weaver, Jordan and Jones, Jake and Singh, Satbir and Beuth, Jack},
  journal={Additive Manufacturing},
  pages={104133},
  year={2024},
  publisher={Elsevier}
}

@article{zheng2018effects,
  title={Effects of scan speed on vapor plume behavior and spatter generation in laser powder bed fusion additive manufacturing},
  author={Zheng, Hang and Li, Huaixue and Lang, Lihui and Gong, Shuili and Ge, Yulong},
  journal={Journal of Manufacturing Processes},
  volume={36},
  pages={60--67},
  year={2018},
  publisher={Elsevier}
}

@article{khairallah2014mesoscopic,
  title={Mesoscopic simulation model of selective laser melting of stainless steel powder},
  author={Khairallah, Saad A and Anderson, Andy},
  journal={Journal of Materials Processing Technology},
  volume={214},
  number={11},
  pages={2627--2636},
  year={2014},
  publisher={Elsevier}
}

@article{son2023design,
  title={Design and Implementation of Health and Risk Level Assessment for Socially Disadvantaged Patients},
  author={Son, Mingeun and Lee, Gangpyo and Park, Jaeyong and Choi, Min},
  journal={Applied Sciences},
  volume={13},
  number={21},
  pages={12014},
  year={2023},
  publisher={MDPI}
}

@article{safieh2024using,
  title={Using Multiple Machine Learning Models to Predict the Strength of UHPC Mixes with Various FA Percentages},
  author={Safieh, Hussam and Hawileh, Rami A and Assad, Maha and Hajjar, Rawan and Shaw, Sayan Kumar and Abdalla, Jamal},
  journal={Infrastructures},
  volume={9},
  number={6},
  pages={92},
  year={2024},
  publisher={MDPI}
}

@article{kim2024road,
  title={A road defect detection system using smartphones},
  author={Kim, Gyulim and Kim, Seungku},
  journal={Sensors},
  volume={24},
  number={7},
  pages={2099},
  year={2024},
  publisher={MDPI}
}

@article{cerne2001coupling,
  title={Coupling of the interface tracking and the two-fluid models for the simulation of incompressible two-phase flow},
  author={Cerne, Gregor and Petelin, Stojan and Tiselj, Iztok},
  journal={Journal of computational physics},
  volume={171},
  number={2},
  pages={776--804},
  year={2001},
  publisher={Elsevier}
}

@article{saldi2012marangoni,
  title={Marangoni driven free surface flows in liquid weld pools},
  author={Saldi, Zaki Saptari},
  journal={Delft University of Technology, The Netherlands},
  year={2012},
  publisher={Citeseer}
}

@article{esmaeilizadeh2019effect,
  title={On the effect of spatter particles distribution on the quality of Hastelloy X parts made by laser powder-bed fusion additive manufacturing},
  author={Esmaeilizadeh, Reza and Ali, Usman and Keshavarzkermani, Ali and Mahmoodkhani, Yahya and Marzbanrad, Ehsan and Toyserkani, Ehsan},
  journal={Journal of Manufacturing Processes},
  volume={37},
  pages={11--20},
  year={2019},
  publisher={Elsevier}
}

@article{khairallah2016laser,
  title={Laser powder-bed fusion additive manufacturing: Physics of complex melt flow and formation mechanisms of pores, spatter, and denudation zones},
  author={Khairallah, Saad A and Anderson, Andrew T and Rubenchik, Alexander and King, Wayne E},
  journal={Acta Materialia},
  volume={108},
  pages={36--45},
  year={2016},
  publisher={Elsevier}
}

@article{lundberg2017unified,
  title={A unified approach to interpreting model predictions},
  author={Lundberg, Scott},
  journal={arXiv preprint arXiv:1705.07874},
  year={2017}
}

@article{cheng2019computational,
  title={Computational investigation of melt pool process dynamics and pore formation in laser powder bed fusion},
  author={Cheng, Bo and Loeber, Lukas and Willeck, Hannes and Hartel, Udo and Tuffile, Charles},
  journal={Journal of Materials Engineering and Performance},
  volume={28},
  pages={6565--6578},
  year={2019},
  publisher={Springer}
}

@article{cook2020simulation,
  title={Simulation of melt pool behaviour during additive manufacturing: Underlying physics and progress},
  author={Cook, Peter S and Murphy, Anthony B},
  journal={Additive Manufacturing},
  volume={31},
  pages={100909},
  year={2020},
  publisher={Elsevier}
}

@article{klassen2014evaporation,
  title={Evaporation model for beam based additive manufacturing using free surface lattice Boltzmann methods},
  author={Klassen, Alexander and Scharowsky, Thorsten and K{\"o}rner, Carolin},
  journal={Journal of Physics D: Applied Physics},
  volume={47},
  number={27},
  pages={275303},
  year={2014},
  publisher={IOP Publishing}
}

@article{heeling2017melt,
  title={Melt pool simulation for the evaluation of process parameters in selective laser melting},
  author={Heeling, Thorsten and Cloots, Michael and Wegener, Konrad},
  journal={Additive Manufacturing},
  volume={14},
  pages={116--125},
  year={2017},
  publisher={Elsevier}
}

@article{ki2002modeling,
  title={Modeling of laser keyhole welding: Part I. Mathematical modeling, numerical methodology, role of recoil pressure, multiple reflections, and free surface evolution},
  author={Ki, Hyungson and Mazumder, Jyoti and Mohanty, Pravansu S},
  journal={Metallurgical and materials transactions A},
  volume={33},
  pages={1817--1830},
  year={2002},
  publisher={Springer}
}

@article{cao2020mesoscopic,
  title={Mesoscopic-scale simulation of pore evolution during laser powder bed fusion process},
  author={Cao, Liu},
  journal={Computational Materials Science},
  volume={179},
  pages={109686},
  year={2020},
  publisher={Elsevier}
}

@article{bobkov2008thermophysical,
  title={Thermophysical properties of materials for nuclear engineering: a tutorial and collection of data},
  author={Bobkov, V and Fokin, L and Petrov, E and Popov, V and Rumiantsev, V and Savvatimsky, A},
  journal={IAEA, Vienna},
  pages={18--21},
  year={2008}
}

@article{farimani2024fast,
  title={Fast Water Desalination with a Graphene--MoS2 Nanoporous Heterostructure},
  author={Farimani, Omid Barati and Cao, Zhonglin and Farimani, Amir Barati},
  journal={ACS Applied Materials \& Interfaces},
  volume={16},
  number={22},
  pages={29355},
  year={2024},
  publisher={American Chemical Society}
}

@article{snow2023observation,
  title={Observation of spatter-induced stochastic lack-of-fusion in laser powder bed fusion using in situ process monitoring},
  author={Snow, Zackary and Scime, Luke and Ziabari, Amirkoushyar and Fisher, Brian and Paquit, Vincent},
  journal={Additive Manufacturing},
  volume={61},
  pages={103298},
  year={2023},
  publisher={Elsevier}
}

@article{taherkhani2021development,
  title={Development of a defect-detection platform using photodiode signals collected from the melt pool of laser powder-bed fusion},
  author={Taherkhani, Katayoon and Sheydaeian, Esmat and Eischer, Christopher and Otto, Martin and Toyserkani, Ehsan},
  journal={Additive Manufacturing},
  volume={46},
  pages={102152},
  year={2021},
  publisher={Elsevier}
}

@article{andani2017spatter,
  title={Spatter formation in selective laser melting process using multi-laser technology},
  author={Andani, Mohsen Taheri and Dehghani, Reza and Karamooz-Ravari, Mohammad Reza and Mirzaeifar, Reza and Ni, Jun},
  journal={Materials \& Design},
  volume={131},
  pages={460--469},
  year={2017},
  publisher={Elsevier}
}

@article{wang2021characteristics,
  title={Characteristics of droplet spatter behavior and process-correlated mapping model in laser powder bed fusion},
  author={Wang, Di and Dou, Wenhao and Ou, Yuanhui and Yang, Yongqiang and Tan, Chaolin and Zhang, Yingjie},
  journal={journal of materials research and technology},
  volume={12},
  pages={1051--1064},
  year={2021},
  publisher={Elsevier}
}

@article{liu2015investigation,
  title={Investigation into spatter behavior during selective laser melting of AISI 316L stainless steel powder},
  author={Liu, Yang and Yang, Yongqiang and Mai, Shuzhen and Wang, Di and Song, Changhui},
  journal={Materials \& Design},
  volume={87},
  pages={797--806},
  year={2015},
  publisher={Elsevier}
}

@article{alphonso2023possibility,
  title={On the possibility of doing reduced order, thermo-fluid modelling of laser powder bed fusion (L-PBF)--Assessment of the importance of recoil pressure and surface tension},
  author={Alphonso, WE and Baier, M and Carmignato, S and Hattel, JH and Bayat, M},
  journal={Journal of Manufacturing Processes},
  volume={94},
  pages={564--577},
  year={2023},
  publisher={Elsevier}
}

@article{tian2020numerical,
  title={Numerical analysis of powder bed generation and single track forming for selective laser melting of SS316L stainless steel},
  author={Tian, Yunfu and Yang, Lijun and Zhao, Dejin and Huang, Yiming and Pan, Jiajing},
  journal={Journal of manufacturing processes},
  volume={58},
  pages={964--974},
  year={2020},
  publisher={Elsevier}
}

@article{kloss2011liggghts,
  title={LIGGGHTS--open source discrete element simulations of granular materials based on Lammps},
  author={Kloss, Christoph and Goniva, Christoph},
  journal={Supplemental Proceedings: Materials Fabrication, Properties, Characterization, and Modeling},
  volume={2},
  pages={781--788},
  year={2011},
  publisher={Wiley Online Library}
}

@article{mindlin1953elastic,
  title={Elastic spheres in contact under varying oblique forces},
  author={Mindlin, Raymond D and Deresiewicz, HERBERT},
  year={1953},
  publisher={American Society of Mechanical Engineers}
}

\pagebreak
\appendix
\section{Machine Learning Models}
\subsubsection{\normalfont Random Forest Classifier
(RFClassifier)}
 Random Forest Classifier  is an ensemble learning method that builds numerous decision trees during training and outputs the class that is the most frequent among the individual trees' predictions \cite{breiman2001random}.  Random forests help mitigate the tendency of the decision trees to overfit their training data \cite{safieh2024using}.
\subsubsection{\normalfont  BaggingClassifier}
  BaggingClassifier is a meta-estimator ensemble learning method that trains base classifiers on random subsets of the original dataset and then combines their individual predictions to produce a final prediction. \cite{breiman1996bagging}.  This approach helps to minimize the model's variance as it combines multiple learners' strengths and smooths out their predictions.
 \subsubsection{\normalfont  Support Vector Classifier (SVC)} 
  Support Vector Classifier (SVC) is a powerful, flexible, and robust model for classification tasks \cite{cortes1995support}.   It introduced the concept of decision planes to establish decision boundaries. SVC seeks to identify a decision plane that optimizes the margin among different classes in the training set, making it particularly well-suited for binary classification problems \cite{kim2024road}.
\subsection{\normalfont Ridge Linear Regression (Ridge)}
Ridge regression is a regularized version of linear regression, used to reduce overfitting which can occur when the model is over-specified for a small dataset \cite{Ridge}. This method performs $L_2$ regularization. Ridge regression shrinks the parameters by imposing a constraint on their values, helping to reduce the model complexity and overfitting. 

\subsection{\normalfont Extreme Gradient Boosting classifier (XGBClassifier)}
EXtreme Gradient Boosting classifier (XGBClassifier) implements  gradient boosted decision trees designed for speed and performance \cite{chen2016xgboost}\cite{son2023design}.  It is known for its efficiency at handling large datasets and its effectiveness in improving model performance through boosting. XGBoost also includes regularization to reduce overfitting.
\subsection{\normalfont Light Gradient Boosting Machine (LGBMClassifier)}
Light Gradient Boosting Machine (LGBMClassifier) is a gradient boosting framework that utilizes tree-based learning algorithms. It is developed for distributed and efficient training, particularly on large datasets \cite{ke2017lightgbm}. LGBM is known for its high efficiency, handling large data, and lower memory usage. Its efficiency comes from histogram-based optimizations in the tree learning process. LGBMClassifier utilizes tree-based learning algorithms because they offer a balance of accuracy, efficiency, and flexibility, making them highly suitable for structured data and large-scale classification problems. The combination of gradient boosting and optimized tree construction techniques further enhances its performance and scalability.

\begin{figure*}[h!]
\begin{center}
\includegraphics[width=1\linewidth]{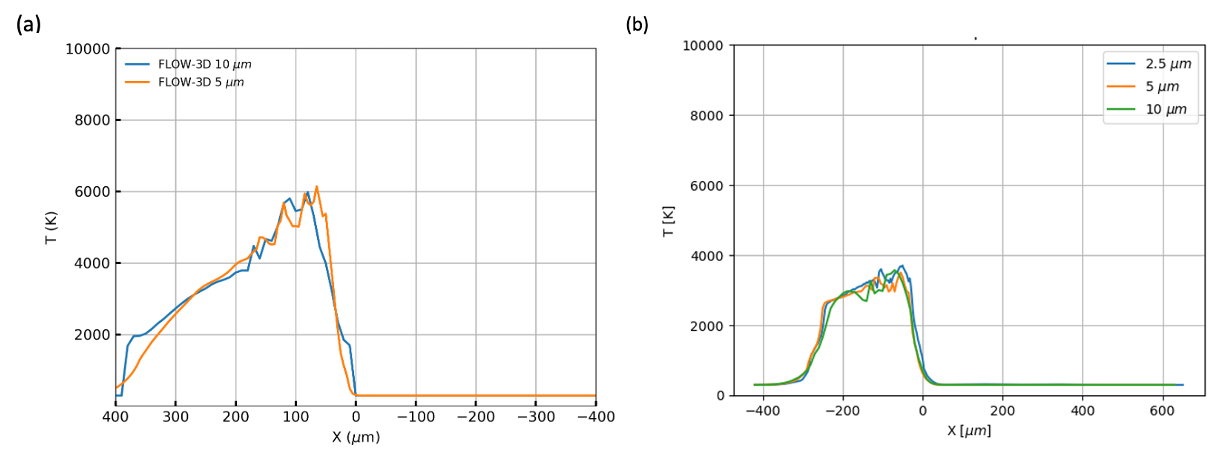}
\end{center}
    \caption{Mesh dependence studies on the temperature profile (a) FLOW-3D (b) OpenFOAM }
\label{fig:mesh_study}
\end{figure*}

\begin{figure*}[htbp!]
\begin{center}
\includegraphics[width=0.8\linewidth]{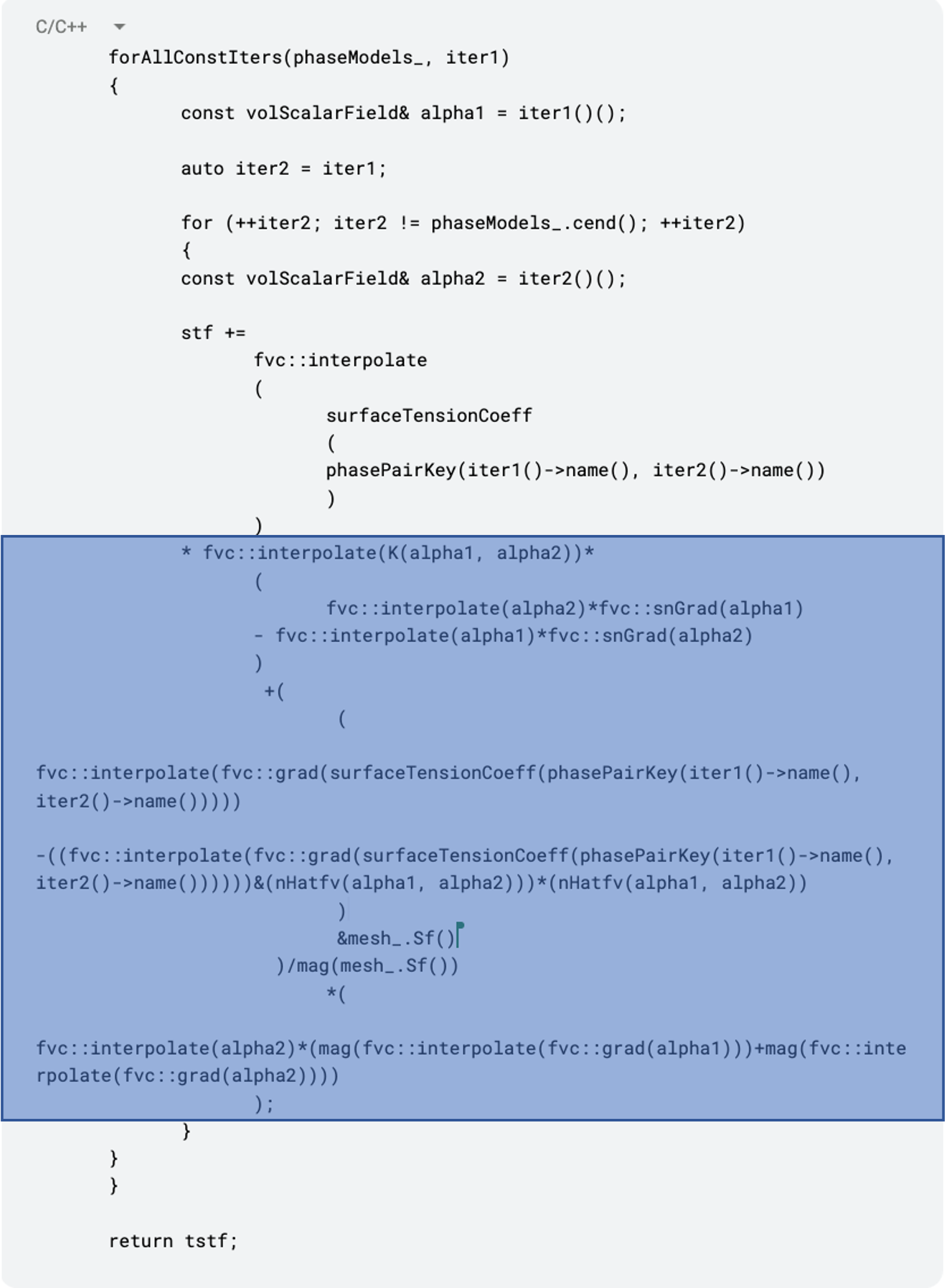}
\end{center}
    \caption{Codeblock with highlighted block to indicated added tangential surface tension to the phaseSystem.C file of the uncompiled solver source code) }
\label{fig:codeblock}
\end{figure*}

\end{document}